\documentclass[sigconf,authorversion]{acmart}

\usepackage{url}
\usepackage{amsmath}
\usepackage{mathtools}
\usepackage{graphicx}
\usepackage[printonlyused]{acronym}
\usepackage{tabularx}
\usepackage{multirow}
\usepackage{booktabs}
\usepackage{graphics}
\usepackage{multicol}
\usepackage{subcaption}
\usepackage{float}
\usepackage{xcolor}
\usepackage{colortbl}

\DeclareMathOperator{\similarity}{sim}

\DeclareMathOperator{\t2i}{t2i}
\DeclareMathOperator{\i2t}{i2t}
\DeclareMathOperator{\infonceloss}{InfoNCE}
\DeclareMathOperator{\selfsuper}{Self-Distil}

\AtBeginDocument{%
  }

\copyrightyear{2024} 
\acmYear{2024} 
\setcopyright{rightsretained} 
\acmConference[SIGSPATIAL '24]{The 32nd ACM International Conference on Advances in Geographic Information Systems}{October 29-November 1, 2024}{Atlanta, GA, USA}
\acmBooktitle{The 32nd ACM International Conference on Advances in Geographic Information Systems (SIGSPATIAL '24), October 29-November 1, 2024, Atlanta, GA, USA}
\acmDOI{10.1145/3678717.3691318}
\acmISBN{979-8-4007-1107-7/24/10}

\begin{document}

\title{Multilingual Vision-Language Pre-training for the Remote Sensing Domain}


\author{João Daniel Silva}
\affiliation{%
  \institution{INESC-ID}
  \institution{Instituto Superior Técnico, University of Lisbon}
  \city{Lisbon}
  \country{Portugal}}
\email{joao.daniel.silva@tecnico.ulisboa.pt}

\author{João Magalhães}
\affiliation{%
  \institution{NOVA-LINCS}
  \institution{Faculty of Science and Technology, NOVA University}
  \city{Lisbon}
  \country{Portugal}}
\email{jmag@fct.unl.pt}

\author{Devis Tuia}
\affiliation{%
  \institution{EPFL ENAC IIE ECEO}
  \institution{École Polytechnique Fédérale de Lausanne}
  \city{Sion}
  \country{Switzerland}}
\email{devis.tuia@epfl.ch}

\author{Bruno Martins}
\affiliation{%
  \institution{INESC-ID \& LUMLIS {\footnotesize(Lisbon ELLIS Unit)}
}
  \institution{Instituto Superior Técnico, University of Lisbon}
  \city{Lisbon}
  \country{Portugal}}
\email{bruno.g.martins@tecnico.ulisboa.pt}
\renewcommand{\shortauthors}{Silva et al.}

\begin{abstract}
Methods based on Contrastive Language-Image Pre-training (CLIP) are nowadays extensively used in support of vision-and-language tasks involving remote sensing data, such as cross-modal retrieval. The adaptation of CLIP to this specific domain has relied on model fine-tuning with the standard contrastive objective, using existing human-labeled image-caption datasets, or using synthetic data corresponding to image-caption pairs derived from other annotations over remote sensing images (e.g., object classes). The use of different pre-training mechanisms has received less attention, and only a few exceptions have considered multilingual inputs. This work proposes a novel vision-and-language model for the remote sensing domain, exploring the fine-tuning of a multilingual CLIP model and testing the use of a self-supervised method based on aligning local and global representations from individual input images, together with the standard CLIP objective. Model training relied on assembling pre-existing datasets of remote sensing images paired with English captions, followed by the use of automated machine translation into nine additional languages. We show that translated data is indeed helpful, e.g. improving performance also on English. Our resulting model, which we named \textbf{R}emote \textbf{S}ensing \textbf{M}ultilingual \textbf{CLIP} (\textbf{RS-M-CLIP}), obtains state-of-the-art results in a variety of vision-and-language tasks, including cross-modal and multilingual image-text retrieval, or zero-shot image classification.
\end{abstract}




\maketitle

\section{Introduction}

Contrastive Language-Image Pre-training (CLIP) was an important contribution in the area of multimodal machine learning~\cite{radford2021Learninga}. In brief, CLIP models are based on an architecture that features two separate encoders, respectively for visual and textual inputs, trained with a contrastive objective that promotes similar semantics and concepts, in both modalities, to be aligned in the embedding space.

CLIP models are nowadays extensively used in a variety of vision-and-language tasks. These include zero-shot classification leveraging class descriptions~\cite{radford2021Learninga,dong2022clip,cherti2023reproducible}, cross-modal retrieval~\cite{radford2021Learninga,remoteclip,Baldrati_2022_CVPR,wang2022medclip}, or object detection/segmentation from text descriptions~\cite{zhou2021extract,wang2023sclip,hajimiri2024pay,zhou2023zegclip,ding2023maskclip}. CLIP is also commonly used as a vision encoder within methods that leverage Large Language Models (LLMs) for supporting multimodal interactions~\cite{liu2023visual,koh2023Grounding,zhao2023mmicl,chen2023shikra,chen2023minigpt}. 

The successful application to different tasks has also led to the adaptation of CLIP for specific domains, such as medical imaging~\cite{wang2022medclip,eslami2023pubmedclip} or remote sensing~\cite{remoteclip,silva2024large,li2023rsclip,zhang2024rs5m,pal2021Fine,rahhal2022multilanguage}. This last case is motivated by the increasing amount of Earth Observation (EO) images being collected through satellites and other aerial vehicles, and also by the interest in methods that can support natural language interactions with large EO image repositories. CLIP models are a natural choice to support vision-and-language tasks over these data, given their good performance in tasks such as cross-modal retrieval, while at the same time avoiding highly specialized mechanisms to extract effective representations from the images.

It should nonetheless be noted that CLIP models typically involve a large number of parameters, thus requiring large datasets for training. For context, the CLIP model associated to the original OpenAI proposal was trained with 400 million image-caption pairs, and other open-sourced variants used billion-scale datasets crawled from the Web. In comparison, despite the high availability of remote sensing imagery, the size of current image-text datasets specific to this domain is much smaller. Aggregating several popular remote sensing  datasets leads to a total of approximately 200k image-text pairs (i.e., considering NWPU-Captions~\cite{cheng2022NWPUCaptions}, RSITMD~\cite{yuan2022exploring}, RSICD~\cite{lu2017exploring}, UCM-Captions~\cite{qu2016Deep}, and Sydney-Captions~\cite{qu2016Deep}). Some studies have advanced pipelines to scale the number of image-caption pairs~\cite{remoteclip}, for instance automatically transforming the annotations associated to other datasets originally designed for semantic segmentation or object detection. The current state-of-the-art model for the remote sensing domain corresponds to  RemoteCLIP~\cite{remoteclip}, which was developed by fine-tuning a large CLIP model with more than 800k synthetic image-caption pairs.

Going beyond the current state-of-the-art, this work proposes a novel vision-and-language model for the remote sensing domain, exploring the fine-tuning of a multilingual CLIP model originally trained on generalist Web data. A large dataset was created by assembling pre-existing remote sensing images paired with English captions, followed by the use of automated machine translation into nine additional languages, as a form of data augmentation~\cite{yuan2023multilingual}. Using these data, the fine-tuning process employed a self-supervised method based on aligning local and global representations from individual input images, together with the standard CLIP objective. 

Through extensive experiments, we show that the use of translated data is indeed helpful, e.g. improving performance on English while enabling multilingual applications. Our resulting model, which we named Remote Sensing Multilingual
CLIP (RS-M-CLIP), obtains state-of-the-art results in a variety of tasks, including cross-modal and multilingual image-text retrieval, or zero-shot image classification. RS-M-CLIP is also capable of producing detailed local features, which hold promise for the application in tasks that require fine-grained localization, such as object detection or semantic segmentation from multilingual descriptions.

Our main contributions can thus be summarized as follows:
\begin{itemize}
    \item A CLIP model for the remote sensing domain that can handle textual inputs in 10 different languages. Both the model and the corresponding source code are publicly available\footnote{\url{https://github.com/DannielSilva/RS-M-CLIP}};
    
    \item Experiments showing that the model obtains state-of-the-art performance on well-known benchmarks, considering tasks such as image-text retrieval (for English and also other languages) or zero-shot image classification. 
\end{itemize}

The rest of this document is organized as follows: Section 2 covers previous related work. Section 3 presents RS-M-CLIP, detailing the training procedure and the data augmentation strategy. Section 4 presents the experimental setup, while Section 5 presents the main results, and Section 6 presents qualitative examples. Finally, Section 7 summarizes our conclusions and discusses future work.

\vspace{-0.15cm}
\section{Related Work}

The current volume of remotely sensed Earth Observation (EO) data motivates the development of vision-and-language methods that can support natural language interactions with large image repositories. Given the success of CLIP for generalist zero-shot image classification and image-text retrieval, several recent studies have discussed the use and adaptation of CLIP for the remote sensing domain. This section surveys relevant previous efforts.

\vspace{-0.15cm}
\subsection{Contrastive Vision-Language Pre-training}\label{rw-clip}

CLIP models~\cite{radford2021Learninga} are usually based on two separate Transformer encoders, respectively for visual and textual inputs. The representations obtained for each modality are trained using a contrastive objective~\cite{oord2018Representation} that promotes similar semantic contents to be close in the resulting embedding space. After training with Web-scale data, CLIP models exhibit strong capabilities for zero-shot image classification and cross-modal retrieval.

Motivated by the strong results, the adaptation of CLIP to specific domains is currently being actively researched. For instance, in the medical imaging domain, radiology images paired to expert medical reports have been used for contrastive training~\cite{wang2022medclip,eslami2023pubmedclip}, resulting in models that can support several clinical downstream tasks.

In the specific domain of remote sensing, several recent efforts are worth of mention. For instance Pal et al.~\cite{pal2021Fine} fine-tuned CLIP with image-caption pairs, observing high-quality results for zero-shot image classification. Rahhal et al.~\cite{rahhal2022multilanguage} fine-tuned CLIP both in mono- and multilingual scenarios, observing good results for cross-modal image-text retrieval. Yuan et al.~\cite{yuan2023parameterefficient} developed a parameter efficient fine-tuning strategy, based on an adapter mechanism, to address image-text retrieval for the remote sensing domain.

Some studies have addressed the problem of the small size of existing remote sensing image-caption datasets, developing different data creation pipelines. For instance Liu et al.~\cite{remoteclip} aggregated information from existing datasets with annotations for semantic segmentation and object detection, using the resulting dataset to train a large CLIP model that achieves superior results in image-text retrieval. Zhang et al.~\cite{zhang2024rs5m} introduced RS5M, i.e. a dataset that collects real instances filtered from publicly available generalist datasets that also contain remote sensing images, and synthetic instances with captions derived from remote sensing image classification datasets. The authors also studied different parameter-efficient approaches to adapt CLIP with the RS5M dataset.

\vspace{-0.15cm}
\subsection{Combining CLIP with Language Decoders}


CLIP vision encoders are commonly used as backbones within recent architectures leveraging Large Language Models (LLMs) for vision-and-language tasks. For instance, LLaVA~\cite{liu2023visual}, MiniGPT-v2~\cite{chen2023minigpt}, or InstructBLIP~\cite{dai2023instructblip} are all examples of instruction fine-tuned models capable of supporting conversation and reasoning over visual inputs, connecting a CLIP vision encoder to a pre-trained LLM, and using supervised learning to train the cross-modal connections. Considering the remote sensing domain, similar efforts include RSGPT~\cite{hu2023rsgpt}, GeoChat~\cite{kuckreja2023geochat}, SkyEyeGPT~\cite{zhan2024skyeyegpt}, or LHRS-Bot~\cite{muhtar2024lhrs}. In previous work, we also tested the use of a CLIP model that was fine-tuned over an aggregated remote sensing dataset, connecting its vision encoder to a pre-trained LLM, so as to support generative tasks such as image captioning~\cite{silva2024large}.

Some previous studies have also advanced training-free strategies for connecting CLIP models to LLMs. For instance Socratic models~\cite{zeng2022socratic} corresponds to a generalist framework in which an LLM is used to process a prompt that integrates information from different models. Under this framework, a task like image captioning can be performed by using CLIP to retrieve different concepts from an input image (e.g., locations or objects that are likely featured in the image), and then prompting an LLM with the retrieved information. LMCap~\cite{ramos2023lmcap} considered a similar strategy, addressing multilingual image captioning through the use of CLIP to retrieve similar captions from a datastore of examples, which are then used to prompt an LLM into generating the target caption. In Appendix~\ref{appendix:captioning}, we present results from a set of experiments using the LMCap method together with our RS-M-CLIP model.

\vspace{-0.15cm}
\subsection{Alternatives to Contrastive Pre-training}

Recent research has addressed the problem of how to improve vision-and-language model performance, complementing CLIP with better data selection strategies and different training objectives. 

Improving the quality of the captions that are paired to images in the training data has been shown to lead to better results, also with better data efficiency~\cite{fan2024improving,vasu2024clip}, while data augmentation has also been observed to be useful, in particular for domains involving smaller datasets~\cite{li2024scaling}. NegCLIP~\cite{yuksekgonul2022and} proposed the use of hard negatives within the CLIP training batches, while CE-CLIP~\cite{zhang2023contrasting} introduced two loss components: one for maximizing the difference between real and hard negative captions, and another making the similarity between a real image-caption pair, and a negative instance, be larger than a threshold. Similar strategies hold significant potential when adapting CLIP to the remote sensing domain.

The combination of self-supervised learning objectives, together with the original contrastive loss, has also been explored. SLIP~\cite{mu2022slip} included the SimCLR~\cite{chen2020Simple} objective, while SILC~\cite{naeem2023silc} introduced a self-distillation objective and observed improvements in zero-shot generalist image classification, cross-modal retrieval, and zero-shot semantic segmentation. EVA~\cite{fang2023eva} addressed the limits of scaling vision models based on masked image modeling, proposing to use CLIP features as the prediction targets. The resulting model obtained state-of-the-art results in a broad range of vision tasks, and it is also commonly used as an encoder for vision and language LLMs~\cite{dai2023instructblip,chen2023minigpt}. In our work, we also considered the combination of contrastive learning with a self-distillation objective, when training a CLIP model, aiming to improve the capabilities of the corresponding vision encoder for the remote sensing domain.

\vspace{-0.15cm}
\subsection{Detailed Localization with CLIP}

Given its zero-shot ability to match images to open textual descriptions, CLIP has also been used within models that feature decoders specific to tasks such as open-vocabulary object detection or semantic segmentation. Approaches like Zegformer~\cite{ding2022decoupling} or zzseg~\cite{xu2022simple} are fully-supervised, 
but training-free models/approaches that explore the feature maps obtained directly from CLIP have also been proposed. Specifically for the remote sensing domain, Chen et al.~\cite{chen2023toward} extended a U-Net-like structure to align its representations with CLIP, allowing for open-vocabulary segmentation. 

In connection to tasks involving detailed localization, the quality of the dense features (i.e., features computed per each image pixel) obtained from CLIP models has also been the subject of recent research. While directly using CLIP has been observed to lack the ability to precisely localize visual concepts within an image, techniques have been developed to slightly modify the architecture of CLIP models when extracting these features, while remaining training-free. For instance MaskCLIP~\cite{zhou2021extract} discards the processing of query and key vectors in the last attention layer of the vision encoder. SCLIP~\cite{wang2023sclip} observes that CLIP learns spatial-invariant visual features, in the sense that local features tend to be invariant to their spatial position in the image, proposing a new self-attention mechanism to obtain spatial covariant features. NACLIP~\cite{hajimiri2024pay} introduces a mechanism that explicitly increases attention values between vicinity patches of the input image, based on the heuristic that adjacent patches often represent the same class.

\vspace{-0.15cm}
\subsection{Cross-Modal Retrieval}

The growing volume of remote sensing imagery has motivated the proposal of specialized cross-modal retrieval methods, e.g. supporting information discovery through text queries over large repositories. 
For instance, GaLR~\cite{yuan2022remote} uses a custom attention mechanism to combine global image features extracted from a CNN, and local features obtained with a graph convolution network. KCR~\cite{mi2022knowledge} and KTIR~\cite{mi2024knowledge} introduce external knowledge from a knowledge graph to enrich text descriptions, and alleviate the information gap between images and text. Instead of specialized architectures for cross-modal retrieval, other studies have instead leveraged CLIP~\cite{radford2021Learninga}, adapting models to the remote sensing domain~\cite{rahhal2022multilanguage,remoteclip,zhang2024rs5m} as detailed in Section~\ref{rw-clip}. For instance, Yuan et al.~\cite{yuan2023parameterefficient} developed a CLIP adapter specifically for the remote sensing image-text retrieval task.

\vspace{-0.15cm}
\section{Remote Sensing Multilingual CLIP}
\begin{figure}[t!]
\centering
\includegraphics[width=\linewidth]{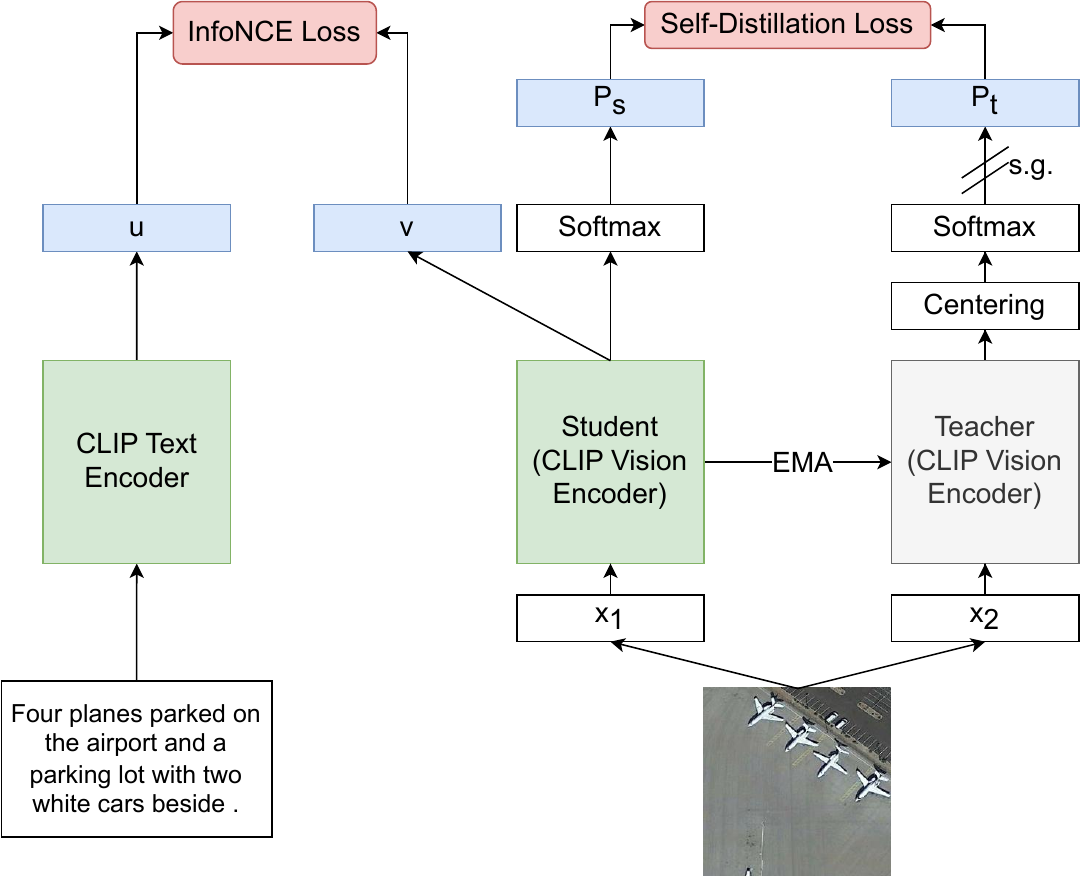}
\caption{
Overview of our method, where self-distillation is combined with contrastive learning to fine-tune a CLIP model using image-caption pairs from remote sensing datasets. Following DINO~\cite{naeem2023silc,caron2021emerging}, local and global image crops are fed into student and teacher networks. Each network produces a probability distribution of size $K$, and the student learns to match the teacher's distribution using a cross-entropy loss, aligning local and global views. The student is updated via gradient updates, while the teacher is updated with an Exponential Moving Average (EMA). The standard CLIP loss for matching image-text pairs is also applied.
} \label{method}
\end{figure}

Our work explored the use of a self-distillation mechanism, together with the conventional contrastive objective from CLIP, aiming at the development of an improved vision-and-language encoder model for the remote sensing domain. Figure~\ref{method} illustrates our proposal.

In brief, the self-distillation mechanism consists of training the vision encoder of CLIP to align local and global views of a given image, so as to improve the handling of fine-grained image details. This is done by passing different views of an image to student and teacher networks based on the CLIP vision encoder. The resulting output is a probability distribution of $K$ dimensions, which the student is trained to match through the cross-entropy loss over the teacher outputs. With this loss, the student parameters are updated with gradient descent, while the teacher parameters are updated with an exponential moving average of the student parameters.

Besides improved training though the self-distillation mechanism, our work also studied the impact of using machine translation to augment the training set of captions paired to remote sensing images. A state-of-the-art LLM for machine translation was used to translate English captions into nine other languages, resulting in a much richer dataset for model training. Given that captions for remote sensing images are often relatively simple sentences, automatic machine translation should not introduce significant errors, thus corresponding to a simple and effective approach with clear immediate benefits (e.g., it becomes possible to support the training of models that can deal with multilingual inputs, while simultaneously aiming at better English performance).

\vspace{-0.15cm}
\subsection{Model Architecture}

The CLIP model consists of a dual Transformer encoder architecture designed to obtain representations for images and texts. The original OpenAI CLIP model cannot process text on other languages besides English, leading us to use a variant that uses XLM-RoBERTa~\cite{conneau2019unsupervised} as the text encoder. For the application of the self-distillation objective, a slight modification over the CLIP vision encoder component is done, where a projector is added on top of the student and teacher. This projector consists of a 3-layer multi-layer perceptron, a $l_2$ normalization operation, and a weight normalized~\cite{salimans2016weight} fully connected layer with $K$ dimensions.

\vspace{-0.15cm}
\subsection{Training Procedure}

An architectural diagram of the proposed training process can be seen in Figure~\ref{method}.
Our method involves joint training with two losses, combining contrastive learning between image and text embeddings (i.e., the standard Information Noise Contrastive Estimation (InfoNCE) loss) together with a self-distillation objective inspired on DINO~\cite{caron2021emerging,naeem2023silc}. We start from a pre-trained generalist CLIP model, using the proposed method to further fine-tune the model to the specific characteristics of remote sensing data.

\vspace{-0.15cm}
\subsubsection{Contrastive Learning}

Contrastive learning can be applied over an unlabeled dataset of paired image-text captions, to maximize the similarity between embeddings of matching pairs~\cite{radford2021Learninga}.  Specifically, for a dataset of $M$ image-caption pairs $\mathcal{D} = \{\mathbf{x}_i, \mathbf{y}_i\}_{i=1}^{M}$, a vision encoder obtains a representation for the input image $f^i_\phi(\mathbf{x}_i) = \mathbf{v}_i \in \mathbb{R}^{m}$, and a text encoder obtains another representation for the caption $f^t_\theta(\mathbf{y}_i) = \mathbf{u}_i \in \mathbb{R}^{m}$. During training, the Information Noise Contrastive Estimation (InfoNCE) loss for both text-to-image $\mathcal{L}_{\t2i}$ and image-to-text $\mathcal{L}_{\i2t}$ matchings is minimized. In a batch of $N$ examples, each image and its corresponding caption are considered a positive pair, while all other combinations of images and captions are considered negative pairs. Given a learnable parameter $\tau$, both losses can be formalized as follows:
\begin{equation} \label{infonceloss_t2i}
    \mathcal{L}_{\t2i} = - \frac{1}{N} \sum^{N}_{i=1} \left( \log \frac{\exp( \similarity(\mathbf{u}_i, \mathbf{v}_i)/\tau)}{\sum^{N}_{j=1}\exp(\similarity(\mathbf{u}_i,\mathbf{v}_i)/\tau)} \right),
\end{equation}
\begin{equation} \label{infonceloss_i2t}
    \mathcal{L}_{\i2t} = - \frac{1}{N} \sum^{N}_{i=1} \left( \log \frac{\exp( \similarity(\mathbf{v}_i,\mathbf{u}_i)/\tau)}{\sum^{N}_{j=1}\exp(\similarity(\mathbf{v}_i,\mathbf{u}_i)/\tau)} \right),
\end{equation}
where the similarity is the cosine similarity given by 
$
    \similarity(\mathbf{a},\mathbf{b}) = \exp(\mathbf{a}\cdot\mathbf{b})/ (\lVert  \mathbf{a}\cdot\mathbf{b} \rVert  \lVert \mathbf{a}\cdot\mathbf{b} \rVert).
$ The final contrastive learning loss is: 
\begin{equation}
    \mathcal{L}_{\infonceloss} = - \frac{1}{2} (\mathcal{L}_{\t2i} + \mathcal{L}_{\i2t}).
\end{equation}

\vspace{-0.15cm}
\subsubsection{Self-distillation}

Besides contrastive learning, we also included a self-distillation mechanism in the training procedure, aiming to obtain higher-quality representations from the image encoder. Following DINO~\cite{caron2021emerging}, a student network $g_{\theta_s}$ is trained to match the output given by a teacher network $g_{\theta_t}$, with the networks being parameterized by $\theta_s$ and $\theta_t$, respectively. Given an input image $x$, the output of each network is normalized with a softmax function, and the results can be interpreted as probability distributions over $K$ dimensions, $P_s$ and $P_t$. The objective function consists of minimizing the cross entropy between the two distributions with respect to the parameters of the student network $\theta_s$, while maintaining the parameters of the teacher network $\theta_t$ fixed:
\begin{equation}
    \min_{\theta_s} = - P_t(x) \log (P_s(x)).
\end{equation}

In practice, for each input image, $S$ views are obtained by applying crop and distortion transformations, where two correspond to global views, i.e. $x_1^g$ and $x_2^g$, and the rest are a set of local views, where each one is an image crop of smaller resolution. Then, every view is given to the student network, while only global views are given to the teacher, encouraging similar representations for the local and global views~\cite{caron2021emerging,naeem2023silc}. Thus, the following loss is minimized:
\begin{equation}\label{dino-loss}
    \mathcal{L}_{\selfsuper} = \sum_{x \in \{x_1^g, x_2^g \}} \sum_{\mathclap{\substack{x^\prime \in S\\
                              x^\prime \ne x}}} - P_t(x) \log (P_s(x^\prime)).
\end{equation}

Both the teacher and the student networks share the same architecture, namely the Transformer vision encoder from a pre-trained CLIP model, although the two networks correspond to different sets of weights, respectively $\theta_s$ and $\theta_t$. The student parameters are learned by minimizing Equation~\ref{dino-loss} with stochastic gradient descent, while the parameters of the teacher are updated with an Exponential Moving Average (EMA) of the student weights, with an update rule corresponding to $\theta_t \leftarrow \lambda \theta_t + (1 - \lambda) \theta_s$.

Regarding training stability, and following the DINO setup, collapse of the obtained representations is avoided by performing centering and sharpening over the teacher outputs. This corresponds to subtracting a centering parameter $c$ while dividing the teacher outputs by a temperature parameter $\tau_t$ in the teacher softmax normalization. The centering parameter $c$ is obtained with an exponential moving average of the mean of the features obtained from the teacher network. These operations are performed as follows for each dimension $i \in \{1, \ldots, K\}$:
\begin{equation}
    g_{\theta_t}(x)^{(i)} \leftarrow \frac{\exp{\left( (g_{\theta_t}(x)^{(i)}-c) / \tau_t\right)} }{\sum_{k=1}^K \exp{ \left( (g_{\theta_t}(x)^{(k)}-c) /\tau_t\right) } }.
\end{equation}

The final training loss for the student, including the contrastive learning and self-distillation objectives, is given by:
\begin{equation}
    \mathcal{L} = \frac{1}{2} (\mathcal{L}_{\infonceloss} + \mathcal{L}_{\selfsuper}).
\end{equation}

\vspace{-0.15cm}
\subsection{Multilingual Augmentations for Captions}

The strategy described in the previous section was used for fine-tuning a generalist multilingual CLIP model, using a large dataset built through automatic machine translation.

We specifically used TowerInstruct~\cite{alves2024tower}, i.e. a Large Language Model (LLM) fine-tuned for machine translation, with state-of-the-art performance in several high-resource languages. The model is prompted in a zero-shot manner to translate individual English captions into other languages. We note that captions for remote sensing images are mostly relatively simple sentences, and hence the simple and direct application of this model should suffice at avoiding severe translation errors.

We translated the captions into 9 languages other than English, as supported by TowerInstruct: German, French, Spanish, Chinese, Portuguese, Italian, Russian, Korean, and Dutch.

\section{Experimental Setup}\label{experimental}

This section describes implementation details such as the choice of training hyper-parameters, the image-caption datasets used in the experiments, and the considered evaluation metrics.

\begin{table*}[t!]
\centering
\caption{Description of the different remote sensing image captioning datasets used in our experiments. Cap-5 is an aggregation of five popular remote sensing image captioning datasets, extensively used in previous studies. We note there may be an overlap in the number of Base Classes in the aggregated datasets, corresponding to different class names for the same concept.}
\label{table:datasets}
\begin{tabularx}{\textwidth}{@{}l *5{>{\centering\arraybackslash}X}@{}}
\hline
\multicolumn{1}{c}{\multirow{2}{*}{\textbf{Individual Dataset}}} & \multirow{2}{*}{\textbf{\#Base Classes}} & \multirow{2}{*}{\textbf{\#Images}} & \multirow{2}{*}{\textbf{Image Size}} & \multirow{2}{*}{\textbf{Spatial Resolution}} & \multirow{2}{*}{\textbf{\#Total Captions}} \\
\multicolumn{1}{c}{} &  &  &  &  &  \\ \hline
NWPU-Captions~\cite{cheng2022NWPUCaptions} & 45 & 31,500 & $256\times256$ & approx. 30-0.2m & 157,500 \\
RSICD~\cite{lu2017exploring} & 30 & 10,921 & $224\times224$ & different resolutions & 54,605 \\
Sydney-Captions~\cite{qu2016Deep} & 7 & 613 & $500\times500$ & 0.5m & 3,065 \\
UCM-Captions~\cite{qu2016Deep} & 21 & 2,100 & $256\times256$ & approx. 0.3m & 10,500 \\
RSITMD~\cite{yuan2022exploring} & 32 & 4,743 & $256\times256$ & different resolutions & 23,715 \\ \hline
\multicolumn{1}{c}{\multirow{2}{*}{\textbf{Aggregated Dataset}}} & \multirow{2}{*}{\textbf{\#Base Classes}} & \multirow{2}{*}{\textbf{\#Images}} & \multicolumn{2}{c}{\multirow{2}{*}{\textbf{Datasets Included}}} & \multirow{2}{*}{\textbf{\#Total Captions}} \\
\multicolumn{1}{c}{} &  &  & \multicolumn{2}{c}{} &  \\ \hline
\cellcolor{red!15}Cap-5 & \cellcolor{red!15}135 & \cellcolor{red!15}49,877 & \multicolumn{2}{c}{\cellcolor{red!15}RET-3 + NWPU-Captions, Sydney-Captions} & \cellcolor{red!15}249,385 \\
RET-3 & 83 & 13,634 & \multicolumn{2}{c}{RSICD, UCM-Captions, RSITMD} & 68,170 \\
RemoteCLIP~\cite{remoteclip} & 194 & 165,745 & \multicolumn{2}{c}{RET-3 + DET-10 + SEG-4} & 828,725 \\ \hline
\end{tabularx}
\end{table*}

\vspace{-0.15cm}
\subsection{Implementation Details}

The models considered in our experiments were implemented using the PyTorch~\cite{paszke2019pytorch} library, and we also used the OpenCLIP software library~\cite{ilharco_gabriel_2021_5143773}. The base multilingual CLIP model used in most of our experiments was $\textit{xlm-roberta-base-ViT-B-32}$, with pre-trained weights corresponding to a checkpoint named $\textit{laion5b\_s13b\_b90k}$\footnote{\url{https://huggingface.co/laion/CLIP-ViT-B-32-xlm-roberta-base-laion5B-s13B-b90k}\label{hfxlm}}. Interestingly, the large-scale LAION-5B dataset of images collected from the Web, originally used to train this model, already includes a significant amount of remote sensing imagery~\cite{zhang2024rs5m}.

Regarding training hyper-parameters, the batch size was set to 128, the learning rate was set to 0.00025, and optimization used the AdamW optimizer for 100 epochs, with a warmup of 10 epochs. As mentioned before, in the setup using self-distillation training, only the student parameters are updated with stochastic gradient descent, while the teacher parameters are updated with an Exponential Moving Average (EMA) from the student weights. The momentum of the EMA update of the teacher is set to $0.996$, the dimensionality of the output probability distribution of the networks is set to $K=65536$, and the hidden dimensionality is set to $2048$, as per the default parameters in DINO. For the contrastive learning objective, similarities are computed between the embeddings of the caption and the first global view. Regarding image transformations, we followed the ones proposed by DINO, which include color jittering, Gaussian blur, and solarization, when obtaining the multiple image crops. As previously mentioned, a total of $S$ views are generated for each input image, consisting of $2$ global views created by randomly cropping the image to $224\times224$ pixels, and $8$ local views with a crop size of $96\times96$ pixels. Due to the positional embeddings of the CLIP vision encoder, images can only be passed as input with the size of $224\times224$ pixels, so local views are scaled up again using bicubic interpolation.

For generating translations of the captions, TowerInstruct-13B-v0.1~\cite{alves2024tower} was prompted in a zero-shot manner with the following instruction: \texttt{ Translate the following text from English into \{language\}.\textbackslash nEnglish: \{caption\}\textbackslash n\{language\}:}. 

\vspace{-0.15cm}
\subsection{Datasets}

Similarly to RemoteCLIP~\cite{remoteclip}, we combined different remote sensing image captioning datasets within our experiments, using the original splits of training, validation, and testing instances. These datasets are described next, and also summarized in Table~\ref{table:datasets}.
\begin{itemize}
    \item RSCID~\cite{lu2017exploring} features a large set of images collected from different sources. The images were manually annotated with short captions, but many of the captions are duplicated to ensure 5 captions per image;
    \item UCM-Captions and Sydney-Captions~\cite{qu2016Deep} are based on scene classification datasets that were repurposed for image captioning by manual annotation. Although extensively used, these datasets are small and have limitations in the fact that the captions correspond to very simple sentences that are highly similar between themselves;
    \item RSITMD~\cite{yuan2022exploring} was proposed to address limitations in other datasets, featuring annotations with fine-grained information, and with a high overall difference between sentences. 
    \item NWPU-Captions~\cite{cheng2022NWPUCaptions} is currently the largest dataset of image and text pairs in the remote sensing domain. Each image has 5 manually annotated captions associated with it, and the images span over a large set of classes describing different land cover and land use types.
\end{itemize}

We aggregated the aforementioned datasets and subsequently reference them as a group through the name \textbf{Cap-5}. For comparison, statistics about the RemoteCLIP dataset~\cite{remoteclip} are also included in Table~\ref{table:datasets}. This other dataset consists of an aggregation of a high volume of images from datasets associated to different tasks (i.e., datasets for object detection (DET-10) and semantic segmentation (SEG-4) tasks), with synthetic captions generated from the available annotations. We refer the reader to the RemoteCLIP paper~\cite{remoteclip} for information regarding the data creation pipeline. RET-3 is also included, which corresponds to the aggregation of the RSICD, UCM-Captions, and RSITMD datasets.

\subsection{Evaluation Metrics}\label{evaluation}
For image-text retrieval, following previous work~\cite{remoteclip, mi2024knowledge}, we report the retrieval recall at top-1 (R@1), top-5 (R@5), and top-10 (R@10) positions, together with the mean recall (mR). R@K simply measures the ratio of queries that successfully retrieve the ground truth as one of the first $K$ results. 

For the zero-shot classification task, we follow~\cite{remoteclip} and use a single template of the form \texttt{ a satellite photo of \{class name\}} to prompt the text encoder of CLIP to generate an embedding for each class of the dataset. For each image, the predicted class is the one with the most similar embedding to the embedding of the image. In the case of remote sensing scene classification datasets that do not have pre-established splits for training and testing, we set a seed value ($42$), randomly shuffle a list of the image filenames for each class, and split the data leaving the first 80\% of the elements for training and the remaining 20\% for testing.

\begin{table*}[t!]
\centering
\caption{Image-text retrieval results for the remote sensing domain. Models with names in \textbf{bold} were fine-tuned by us, and results obtained through our own experiments, instead of collected from previous publications, are marked with~\dag.
}
\label{table:clip-objective}
\resizebox{\textwidth}{!}{
\begin{tabular}{clccccccccc}
\hline
\multirow{2}{*}{\textbf{Dataset}} & \multicolumn{1}{c}{\multirow{2}{*}{\textbf{Method}}} & \multirow{2}{*}{\textbf{Vision Backbone}} & \multirow{2}{*}{\textbf{Finetuning Data}} & \multicolumn{3}{c}{\textbf{Image-Text Retrieval}} & \multicolumn{3}{c}{\textbf{Text-Image Retrieval}} & \multirow{2}{*}{\textbf{\begin{tabular}[c]{@{}c@{}}Mean\\ Recall\end{tabular}}} \\ \cline{5-10}
 & \multicolumn{1}{c}{} &  &  & \textbf{R@1} & \textbf{R@5} & \textbf{R@10} & \textbf{R@1} & \textbf{R@5} & \textbf{R@10} &  \\ \hline
\multicolumn{1}{l}{\multirow{18}{*}{\rotatebox{90}{RSICD}}} & AMFMN~\cite{hoxha2020toward} & ResNet50 & RSICD & 5.39 & 15.08 & 23.40 & 4.90 & 18.28 & 31.44 & 16.42 \\
\multicolumn{1}{l}{} & CMFM-Net~\cite{yu2022text} & ResNet18 & RSICD & 5.40 & 18.66 & 28.55 & 5.31 & 18.57 & 30.03 & 17.75 \\
\multicolumn{1}{l}{} & GaLR~\cite{yuan2022remote} & ResNet18 & RSICD & 6.59 & 19.85 & 31.04 & 4.69 & 19.48 & 32.13 & 18.96 \\
\multicolumn{1}{l}{} & KCR~\cite{mi2022knowledge} & ResNet18 & RSICD & 4.76 & 18.59 & 27.20 & 5.84 & 22.31 & 36.12 & 19.14 \\
\multicolumn{1}{l}{} & CLIP~\cite{radford2021Learninga}\dag & ViT-B & Zero-shot & 4.58 & 14.55 & 23.70 & 5.80 & 16.85 & 28.23 & 15.62 \\
\multicolumn{1}{l}{} & CLIP~\cite{radford2021Learninga}\dag & ViT-L & Zero-shot & 6.04 & 17.48 & 27.54 & 5.03 & 19.03 & 30.25 & 17.56 \\
\multicolumn{1}{l}{} & Rahhal et al.~\cite{rahhal2022multilanguage} & ViT-B & RSICD & 10.70 & 29.64 & 41.53 & 9.14 & 28.96 & 44.59 & 27.43 \\
\multicolumn{1}{l}{} & CLIP-RSICD~\cite{pal2021Fine}\dag & ViT-B & RSICD & 14.09 & 30.10 & 43.64 & 11.16 & 33.25 & 48.91 & 30.19 \\
\multicolumn{1}{l}{} & GeoRSCLIP~\cite{zhang2024rs5m} & ViT-B & RS5M & 11.53 & 28.55 & 39.16 & 9.52 & 27.37 & 40.99 & 26.18 \\
\multicolumn{1}{l}{} & RemoteCLIP~\cite{remoteclip} & ViT-B & RET-3 + DET-10 + SEG-4 & 17.02 & 37.97 & 51.51 & 13.71 & 37.11 & 54.25 & 35.26 \\
\multicolumn{1}{l}{} & RemoteCLIP~\cite{remoteclip} & ViT-L & RET-3 + DET-10 + SEG-4 & 18.39 & 37.42 & 51.05 & 14.73 & 39.93 & 56.58 & 36.35 \\
\multicolumn{1}{l}{} & KTIR~\cite{mi2024knowledge} & VIT-B & RSICD & 26.08 & 49.77 & 62.49 & 20.55 & 48.67 & 63.70 & 45.21 \\
\multicolumn{1}{l}{} & \cellcolor{lightgray!50}\textbf{CLIP/Cap-5}\dag & \cellcolor{lightgray!50}ViT-L & \cellcolor{lightgray!50}Cap-5 & \cellcolor{lightgray!50}17.02 & \cellcolor{lightgray!50}33.94 & \cellcolor{lightgray!50}47.76 & \cellcolor{lightgray!50}13.83 & \cellcolor{lightgray!50}39.07 & \cellcolor{lightgray!50}56.05 & \cellcolor{lightgray!50}34.61 \\ \cline{2-11} 
\multicolumn{1}{l}{} & \multicolumn{10}{c}{\textbf{Multilingual Backbones}} \\ \cline{2-11} 
\multicolumn{1}{l}{} & CLIP-XLM-RoBERTa\footref{hfxlm}\dag & ViT-B & Zero-shot & 7.41 & 16.83 & 25.62 & 6.26 & 22.07 & 33.93 & 18.69 \\
\multicolumn{1}{l}{} & \textbf{CLIP-XLM-RoBERTa}\dag & ViT-B & Cap-5 / English only & 27.17 & 50.69 & 64.04 & 20.79 & 48.58 & 62.10 & 45.56 \\
\multicolumn{1}{l}{} & \textbf{RS-M-CLIP}\dag & ViT-B & Cap-5 / English only & 58.55 & 69.08 & 73.65 & 43.17 & 58.79 & 66.57 & 61.64 \\
\multicolumn{1}{l}{} & \cellcolor{red!15}\textbf{RS-M-CLIP}\dag & \cellcolor{red!15}ViT-B & \cellcolor{red!15}Cap-5 / 1 Translation & \cellcolor{red!15}\textbf{60.48} & \cellcolor{red!15}\textbf{72.10} & \cellcolor{red!15}\textbf{75.93} & \cellcolor{red!15}\textbf{44.37} & \cellcolor{red!15}\textbf{60.35} & \cellcolor{red!15}\textbf{69.02} & \cellcolor{red!15}\textbf{63.71} \\ \hline
\multirow{15}{*}{\rotatebox{90}{UCM}} & AMFMN~\cite{hoxha2020toward} & ResNet50 & UCM & 16.67 & 45.71 & 68.57 & 12.86 & 53.24 & 79.43 & 46.08 \\
 & KCR~\cite{mi2022knowledge} & ResNet18 & UCM & 11.90 & 48.57 & 71.43 & 17.24 & 56.95 & 81.14 & 47.87 \\
 & CLIP~\cite{radford2021Learninga}\dag & ViT-B & Zero-shot & 8.10 & 34.29 & 53.33 & 8.67 & 36.48 & 60.57 & 33.57 \\
 & CLIP~\cite{radford2021Learninga}\dag & ViT-L & Zero-shot & 11.91 & 43.33 & 65.24 & 10.76 & 46.00 & 73.33 & 41.76 \\
 & Rahhal et al.~\cite{rahhal2022multilanguage}\dag & ViT-B & UCM & 19.04 & 53.33 & 77.61 & 19.33 & 64.00 & 91.42 & 54.12 \\
 & CLIP-RSICD~\cite{pal2021Fine}\dag & ViT-B & RSICD & 15.71 & 50.00 & 82.38 & 13.81 & 57.05 & 91.24 & 51.70 \\
 & RemoteCLIP~\cite{remoteclip} & ViT-B & RET-3 + DET-10 + SEG-4 & 20.48 & 59.85 & 83.33 & 18.67 & 61.52 & 94.29 & 56.36 \\
 & RemoteCLIP~\cite{remoteclip} & ViT-L & RET-3 + DET-10 + SEG-4 & 19.05 & 54.29 & 80.95 & 17.71 & 62.19 & 93.90 & 54.68 \\
 & KTIR~\cite{mi2022knowledge} & ViT-B & UCM & 21.42 & \textbf{64.29} & \textbf{87.14} & \textbf{19.81} & \textbf{64.57} & \textbf{95.33} & \textbf{58.76} \\ 
 & \cellcolor{lightgray!50}\textbf{CLIP/Cap-5}\dag & \cellcolor{lightgray!50}ViT-L & \cellcolor{lightgray!50}Cap-5 & \cellcolor{lightgray!50}\textbf{21.91} & \cellcolor{lightgray!50}59.05 & \cellcolor{lightgray!50}81.91 & \cellcolor{lightgray!50}16.29 & \cellcolor{lightgray!50}60.57 & \cellcolor{lightgray!50}94.76 & \cellcolor{lightgray!50}55.75 \\
 \cline{2-11} 
 & \multicolumn{10}{c}{\textbf{Multilingual Backbones}} \\ \cline{2-11} 
 & CLIP-XLM-RoBERTa\footref{hfxlm}\dag & ViT-B & Zero-shot & 9.52 & 35.24 & 52.38 & 11.62 & 43.71 & 66.38 & 36.48 \\
 & \textbf{CLIP-XLM-RoBERTa}\dag & ViT-B & Cap-5 / English Only & 18.10 & 60.95 & 83.33 & 17.14 & 59.33 & 91.91 & 55.13 \\
 & \textbf{RS-M-CLIP}\dag & ViT-B & Cap-5 / English only& 14.76 & 52.86 & 75.24 & 14.57 & 52.57 & 74.29 & 47.38 \\
 & \cellcolor{red!15}\textbf{RS-M-CLIP}\dag & \cellcolor{red!15}ViT-B & \cellcolor{red!15}Cap-5 / 1 Translation & \cellcolor{red!15}14.29 & \cellcolor{red!15}50.00 & \cellcolor{red!15}77.14 & \cellcolor{red!15}16.29 & \cellcolor{red!15}55.33 & \cellcolor{red!15}83.91 & \cellcolor{red!15}49.49 \\ \hline
\multirow{17}{*}{\rotatebox{90}{RSITMD}} & AMFMN~\cite{hoxha2020toward} & ResNet50 & RSITMD & 10.63 & 24.78 & 41.81 & 11.51 & 34.69 & 54.87 & 29.72 \\
 & CMFM-Net~\cite{yu2022text} & ResNet18 & RSITMD & 10.84 & 28.76 & 40.04 & 10.00 & 32.83 & 47.21 & 28.28 \\
 & GaLR~\cite{yuan2022remote} & ResNet18 & RSITMD & 14.82 & 31.64 & 42.48 & 11.15 & 36.68 & 51.68 & 31.41 \\
 & CLIP~\cite{radford2021Learninga}\dag & ViT-B & Zero-shot & 9.74 & 22.57 & 34.51 & 8.72 & 27.88 & 42.88 & 24.38 \\
 & CLIP~\cite{radford2021Learninga}\dag & ViT-L & Zero-shot & 10.18 & 27.66 & 38.50 & 11.46 & 32.52 & 46.99 & 27.88 \\
 & Rahhal et al.~\cite{rahhal2022multilanguage} & ViT-B & RSITMD & 19.69 & 40.26 & 54.42 & 17.61 & 49.73 & 66.59 & 41.38 \\
 & CLIP-RSICD~\cite{pal2021Fine} & ViT-B & RSICD & 25.00 & 46.46 & 61.73 & 19.29 & 51.02 & 68.45 & 45.32 \\
 & GeoRSCLIP~\cite{zhang2024rs5m} & ViT-B & RS5M & 19.03 & 34.51 & 46.46 & 14.16 & 42.39 & 57.52 & 35.68 \\
 & RemoteCLIP~\cite{remoteclip} & ViT-B & RET-3 + DET-10 + SEG-4 & 27.88 & 50.66 & 65.71 & 22.17 & 56.46 & 73.41 & 49.38 \\
 & RemoteCLIP~\cite{remoteclip} & ViT-L & RET-3 + DET-10 + SEG-4 & 28.76 & 52.43 & 63.94 & 23.76 & 59.51 & 74.73 & 50.52 \\
 & KTIR~\cite{mi2024knowledge} & VIT-B & RSITMD & 34.29 & 55.31 & 65.04 & 31.46 & 62.92 & 76.59 & 54.27 \\
 & \cellcolor{lightgray!50}\textbf{CLIP/Cap-5}\dag & \cellcolor{lightgray!50}ViT-L & \cellcolor{lightgray!50}Cap-5 & \cellcolor{lightgray!50}41.37 & \cellcolor{lightgray!50}67.48 & \cellcolor{lightgray!50}79.65 & \cellcolor{lightgray!50}34.12 & \cellcolor{lightgray!50}68.81 & \cellcolor{lightgray!50}82.12 & \cellcolor{lightgray!50}62.26 \\ \cline{2-11} 
 & \multicolumn{10}{c}{\textbf{Multilingual Backbones}} \\ \cline{2-11} 
 & CLIP-XLM-RoBERTa\footref{hfxlm}\dag & ViT-B & Zero-shot & 8.41 & 26.11 & 34.96 & 11.33 & 34.29 & 49.69 & 27.46 \\
 & \textbf{CLIP-XLM-RoBERTa}\dag & ViT-B & Cap-5 / English only & 39.82 & 64.60 & 75.89 & 38.85 & 68.81 & 80.53 & 61.42 \\
 & \textbf{RS-M-CLIP}\dag & ViT-B & Cap-5 / English only & 63.05 & 75.00 & 80.97 & 55.18 & 73.27 & 80.71 & 71.36 \\
 & \cellcolor{red!15}\textbf{RS-M-CLIP}\dag & \cellcolor{red!15}ViT-B & \cellcolor{red!15}Cap-5 / 1 Translation & \cellcolor{red!15}\textbf{67.48} & \cellcolor{red!15}\textbf{77.66} & \cellcolor{red!15}\textbf{82.30} & \cellcolor{red!15}\textbf{59.87} & \cellcolor{red!15}\textbf{76.99} & \cellcolor{red!15}\textbf{83.01} & \cellcolor{red!15}\textbf{74.55} \\ \hline
\end{tabular}
}
\end{table*}

\begin{table*}[t!]
\centering
\caption{Zero-shot image classification in 12 remote sensing datasets. Due to space constraints, CLIP-XLM-RoBERTa is mentioned in the table simply as CLIP-XLM.
}
\label{table:classification}
\resizebox{\textwidth}{!}{
\begin{tabular}{lcccccccccccccc}
\hline
\multicolumn{1}{c}{\textbf{Method}} & \textbf{Backbone} & \textbf{RSI-CB128} & \textbf{RSI-CB256} & \textbf{WHU-earth} & \textbf{EuroSAT} & \textbf{MLRSNet} & \textbf{PatternNet} & \textbf{RESISC45} & \textbf{AID} & \textbf{RSSCN7} & \textbf{OPTIMAL-31} & \textbf{RSC11} & \textbf{WHU-RS19} & \textbf{Average} \\ \hline
CLIP~\cite{radford2021Learninga} & ViT-B & 20.09 & 29.40 & 40.62 & 41.83 & 41.12 & 49.65 & 50.65 & 51.40 & 65.00 & 66.13 & 46.40 & 73.79 & 48.01 \\
RemoteCLIP~\cite{remoteclip} & ViT-B & 25.88 & 42.56 & 63.12 & 33.04 & 57.91 & 58.01 & 66.87 & 86.60 & 73.75 & 79.03 & 57.16 & 94.17 & 61.51 \\ \hline
CLIP~\cite{radford2021Learninga} & ViT-L & 32.69 & 43.65 & 56.46 & 39.19 & 59.09 & 70.15 & 64.16 & 62.70 & 67.68 & 73.92 & 64.26 & 87.86 & 60.15 \\
RemoteCLIP~\cite{remoteclip} & ViT-L & 37.23 & 51.55 & 68.33 & 46.04 & 61.67 & 63.12 & 73.68 & 76.95 & 71.07 & 80.65 & 68.43 & 92.72 & 65.95 \\ \hline
CLIP-XLM & ViT-B & 31.12 & 41.82 & 59.17 & 52.30 & 55.97 & 60.02 & 62.79 & 67.45 & 69.82 & 75.27 & 56.04 & 85.44 & 59.77 \\
\cellcolor{red!15}RS-M-CLIP & \cellcolor{red!15}ViT-B & \cellcolor{red!15}28.27 & \cellcolor{red!15}38.61 & \cellcolor{red!15}63.96 & \cellcolor{red!15}25.85 & \cellcolor{red!15}60.62 & \cellcolor{red!15}50.95 & \cellcolor{red!15}92.62 & \cellcolor{red!15}88.90 & \cellcolor{red!15}60.89 & \cellcolor{red!15}95.16 & \cellcolor{red!15}61.42 & \cellcolor{red!15}90.78 & \cellcolor{red!15}63.17 \\ \hline
\end{tabular}
}
\end{table*}

\begin{table*}[t!]
\centering
\caption{Image-text retrieval results in over the RSICD dataset, for multilingual CLIP models in 10 different languages. 
}
\label{table:multilingual-rsicd}
\resizebox{\textwidth}{!}{
\begin{tabular}{clccccccccc}
\hline
\multirow{2}{*}{\textbf{Language}} & \multicolumn{1}{c}{\multirow{2}{*}{\textbf{Method}}} & \multirow{2}{*}{\textbf{Visual Backbone}} & \multirow{2}{*}{\textbf{Finetune Data}} & \multicolumn{3}{c}{\textbf{Image-Text Retrieval}} & \multicolumn{3}{c}{\textbf{Text-Image Retrieval}} & \multirow{2}{*}{\textbf{\begin{tabular}[c]{@{}c@{}}Mean\\ Recall\end{tabular}}} \\ \cline{5-10}
 & \multicolumn{1}{c}{} &  &  & \textbf{R@1} & \textbf{R@5} & \textbf{R@10} & \textbf{R@1} & \textbf{R@5} & \textbf{R@10} &  \\ \hline
\multirow{3}{*}{English} & \cellcolor{lightgray!50}Rahhal et al. & \cellcolor{lightgray!50}ViT-B & \cellcolor{lightgray!50}RSICD (Multilanguage) & \cellcolor{lightgray!50}11.61 & \cellcolor{lightgray!50}30.10 & \cellcolor{lightgray!50}42.17 & \cellcolor{lightgray!50}9.55 & \cellcolor{lightgray!50}29.27 & \cellcolor{lightgray!50}44.73 & \cellcolor{lightgray!50}27.90 \\
 & CLIP-XLM-RoBERTa & ViT-B & Zero-shot & 7.41 & 16.83 & 25.62 & 6.26 & 22.07 & 33.93 & 18.69 \\
 & \cellcolor{red!15}RS-M-CLIP & \cellcolor{red!15}ViT-B & \cellcolor{red!15}Cap-5 / 1 Translation & \cellcolor{red!15}60.48 & \cellcolor{red!15}72.10 & \cellcolor{red!15}75.93 & \cellcolor{red!15}44.37 & \cellcolor{red!15}60.35 & \cellcolor{red!15}69.02 & \cellcolor{red!15}63.71 \\ \hline
\multirow{3}{*}{French} & \cellcolor{lightgray!50}Rahhal et al. & \cellcolor{lightgray!50}ViT-B & \cellcolor{lightgray!50}RSICD (Multilanguage) & \cellcolor{lightgray!50}8.78 & \cellcolor{lightgray!50}26.89 & \cellcolor{lightgray!50}40.07 & \cellcolor{lightgray!50}8.19 & \cellcolor{lightgray!50}26.34 & \cellcolor{lightgray!50}43.11 & \cellcolor{lightgray!50}25.56 \\
 & CLIP-XLM-RoBERTa & ViT-B & Zero-shot & 6.13 & 14.55 & 24.06 & 5.65 & 19.01 & 30.34 & 16.62 \\
 & \cellcolor{red!15}RS-M-CLIP & \cellcolor{red!15}ViT-B & \cellcolor{red!15}Cap-5 / 1 Translation & \cellcolor{red!15}57.82 & \cellcolor{red!15}71.36 & \cellcolor{red!15}75.76 & \cellcolor{red!15}43.82 & \cellcolor{red!15}59.51 & \cellcolor{red!15}68.66 & \cellcolor{red!15}62.82 \\ \hline
\multirow{3}{*}{Italian} & \cellcolor{lightgray!50}Rahhal et al. & \cellcolor{lightgray!50}ViT-B & \cellcolor{lightgray!50}RSICD (Multilanguage) & \cellcolor{lightgray!50}10.15 & \cellcolor{lightgray!50}26.98 & \cellcolor{lightgray!50}37.69 & \cellcolor{lightgray!50}8.43 & \cellcolor{lightgray!50}26.56 & \cellcolor{lightgray!50}42.08 & \cellcolor{lightgray!50}25.31 \\
 & CLIP-XLM-RoBERTa & ViT-B & Zero-shot & 5.76 & 15.83 & 23.61 & 5.00 & 18.52 & 29.46 & 16.36 \\
 & \cellcolor{red!15}RS-M-CLIP & \cellcolor{red!15}ViT-B & \cellcolor{red!15}Cap-5 / 1 Translation & \cellcolor{red!15}58.65 & \cellcolor{red!15}71.09 & \cellcolor{red!15}75.48 & \cellcolor{red!15}43.44 & \cellcolor{red!15}59.65 & \cellcolor{red!15}68.75 & \cellcolor{red!15}62.84 \\ \hline
\multirow{2}{*}{Portuguese} & CLIP-XLM-RoBERTa & ViT-B & Zero-shot & 5.86 & 15.83 & 22.60 & 4.58 & 18.59 & 29.55 & 16.17 \\
 & RS-M-CLIP & ViT-B & Cap-5 / 1 Translation & 59.01 & 71.46 & 75.30 & 43.84 & 59.91 & 68.23 & 62.96 \\ \hline
\multirow{2}{*}{Spanish} & CLIP-XLM-RoBERTa & ViT-B & Zero-shot & 6.68 & 15.55 & 23.88 & 5.18 & 19.07 & 29.63 & 16.66 \\
 & RS-M-CLIP & ViT-B & Cap-5 / 1 Translation & 59.10 & 71.18 & 75.48 & 43.75 & 59.96 & 68.89 & 63.06 \\ \hline
\multirow{2}{*}{German} & CLIP-XLM-RoBERTa & ViT-B & Zero-shot & 5.31 & 14.27 & 22.78 & 5.53 & 17.86 & 28.03 & 15.63 \\
 & RS-M-CLIP & ViT-B & Cap-5 / 1 Translation & 54.90 & 68.98 & 74.93 & 38.66 & 56.23 & 65.64 & 59.89 \\ \hline
\multirow{2}{*}{Dutch} & CLIP-XLM-RoBERTa & ViT-B & Zero-shot & 5.03 & 15.01 & 23.06 & 4.63 & 18.50 & 29.84 & 16.01 \\
 & RS-M-CLIP & ViT-B & Cap-5 / 1 Translation & 58.74 & 71.73 & 76.49 & 43.86 & 59.63 & 67.65 & 63.02 \\ \hline
\multirow{2}{*}{Korean} & CLIP-XLM-RoBERTa & ViT-B & Zero-shot & 4.30 & 12.53 & 19.12 & 3.84 & 16.03 & 25.34 & 13.53 \\
 & RS-M-CLIP & ViT-B & Cap-5 / 1 Translation & 56.91 & 69.35 & 73.74 & 42.05 & 58.87 & 67.96 & 61.48 \\ \hline
\multirow{2}{*}{Chinese} & CLIP-XLM-RoBERTa & ViT-B & Zero-shot & 5.40 & 15.46 & 23.42 & 4.41 & 16.18 & 26.97 & 15.31 \\
 & RS-M-CLIP & ViT-B & Cap-5 / 1 Translation & 55.63 & 69.44 & 74.66 & 42.09 & 58.98 & 67.36 & 61.36 \\ \hline
\multirow{2}{*}{Russian} & CLIP-XLM-RoBERTa & ViT-B & Zero-shot & 5.86 & 15.37 & 23.79 & 4.96 & 17.69 & 27.94 & 15.94 \\
 & RS-M-CLIP & ViT-B & Cap-5 / 1 Translation & 57.46 & 71.55 & 75.57 & 42.65 & 59.76 & 68.25 & 62.54 \\ \hline
\end{tabular}
}
\end{table*}

\begin{table*}[t!]
\centering
\caption{Image-text retrieval results in over the RSITMD dataset for multilingual CLIP models in 10 different languages. 
}
\label{table:multilingual-rsitmd}
\resizebox{\textwidth}{!}{
\begin{tabular}{clccccccccc}
\hline
\multirow{2}{*}{\textbf{Language}} & \multicolumn{1}{c}{\multirow{2}{*}{\textbf{Method}}} & \multirow{2}{*}{\textbf{Visual Backbone}} & \multirow{2}{*}{\textbf{Finetune Data}} & \multicolumn{3}{c}{\textbf{Image-Text Retrieval}} & \multicolumn{3}{c}{\textbf{Text-Image Retrieval}} & \multirow{2}{*}{\textbf{\begin{tabular}[c]{@{}c@{}}Mean\\ Recall\end{tabular}}} \\ \cline{5-10}
 & \multicolumn{1}{c}{} &  &  & \textbf{R@1} & \textbf{R@5} & \textbf{R@10} & \textbf{R@1} & \textbf{R@5} & \textbf{R@10} &  \\ \hline
\multirow{3}{*}{English} & \cellcolor{lightgray!50}Rahhal et al. & \cellcolor{lightgray!50}ViT-B & \cellcolor{lightgray!50}RSITMD (Multilanguage) & \cellcolor{lightgray!50}22.56 & \cellcolor{lightgray!50}43.80 & \cellcolor{lightgray!50}57.30 & \cellcolor{lightgray!50}19.29 & \cellcolor{lightgray!50}51.37 & \cellcolor{lightgray!50}68.18 & \cellcolor{lightgray!50}43.75 \\
 & CLIP-XLM-RoBERTa & ViT-B & Zero-shot & 8.41 & 25.89 & 34.96 & 11.37 & 34.25 & 49.69 & 27.43 \\
 & \cellcolor{red!15}RS-M-CLIP & \cellcolor{red!15}ViT-B & \cellcolor{red!15}Cap-5 / 1 Translation & \cellcolor{red!15}67.48 & \cellcolor{red!15}77.66 & \cellcolor{red!15}82.30 & \cellcolor{red!15}59.87 & \cellcolor{red!15}76.99 & \cellcolor{red!15}83.01 & \cellcolor{red!15}74.55 \\ \hline
\multirow{3}{*}{French} & \cellcolor{lightgray!50}Rahhal et al. & \cellcolor{lightgray!50}ViT-B & \cellcolor{lightgray!50}RSITMD (Multilanguage) & \cellcolor{lightgray!50}20.35 & \cellcolor{lightgray!50}43.80 & \cellcolor{lightgray!50}58.18 & \cellcolor{lightgray!50}19.29 & \cellcolor{lightgray!50}49.38 & \cellcolor{lightgray!50}65.61 & \cellcolor{lightgray!50}42.77 \\
 & CLIP-XLM-RoBERTa & ViT-B & Zero-shot & 11.06 & 23.67 & 37.39 & 10.66 & 29.78 & 44.60 & 26.20 \\
 & \cellcolor{red!15}RS-M-CLIP & \cellcolor{red!15}ViT-B & \cellcolor{red!15}Cap-5 / 1 Translation & \cellcolor{red!15}66.59 & \cellcolor{red!15}78.10 & \cellcolor{red!15}82.97 & \cellcolor{red!15}58.89 & \cellcolor{red!15}76.02 & \cellcolor{red!15}83.10 & \cellcolor{red!15}74.28 \\ \hline
\multirow{3}{*}{Italian} & \cellcolor{lightgray!50}Rahhal et al. & \cellcolor{lightgray!50}ViT-B & \cellcolor{lightgray!50}RSITMD (Multilanguage) & \cellcolor{lightgray!50}21.90 & \cellcolor{lightgray!50}42.69 & \cellcolor{lightgray!50}56.63 & \cellcolor{lightgray!50}18.18 & \cellcolor{lightgray!50}49.02 & \cellcolor{lightgray!50}64.91 & \cellcolor{lightgray!50}42.22 \\
 & CLIP-XLM-RoBERTa & ViT-B & Zero-shot & 9.96 & 23.01 & 31.20 & 9.96 & 29.56 & 43.14 & 24.47 \\
 & \cellcolor{red!15}RS-M-CLIP & \cellcolor{red!15}ViT-B & \cellcolor{red!15}Cap-5 / 1 Translation & \cellcolor{red!15}68.14 & \cellcolor{red!15}78.10 & \cellcolor{red!15}82.97 & \cellcolor{red!15}58.63 & \cellcolor{red!15}76.02 & \cellcolor{red!15}82.57 & \cellcolor{red!15}74.40 \\ \hline
\multirow{2}{*}{Portuguese} & CLIP-XLM-RoBERTa & ViT-B & Zero-shot & 9.96 & 23.45 & 32.08 & 9.87 & 29.29 & 44.51 & 24.86 \\
 & RS-M-CLIP & ViT-B & Cap-5 / 1 Translation & 67.48 & 77.43 & 82.97 & 59.12 & 76.15 & 82.88 & 74.34 \\ \hline
\multirow{2}{*}{Spanish} & CLIP-XLM-RoBERTa & ViT-B & Zero-shot & 10.62 & 21.46 & 32.52 & 10.75 & 31.11 & 45.22 & 25.28 \\
 & RS-M-CLIP & ViT-B & Cap-5 / 1 Translation & 67.92 & 77.43 & 82.97 & 59.34 & 76.33 & 82.83 & 74.47 \\ \hline
\multirow{2}{*}{German} & CLIP-XLM-RoBERTa & ViT-B & Zero-shot & 8.41 & 22.35 & 32.97 & 9.91 & 28.67 & 43.85 & 24.36 \\
 & RS-M-CLIP & ViT-B & Cap-5 / 1 Translation & 62.39 & 75.22 & 82.74 & 53.01 & 72.21 & 79.78 & 70.89 \\ \hline
\multirow{2}{*}{Dutch} & CLIP-XLM-RoBERTa & ViT-B & Zero-shot & 10.18 & 21.68 & 31.86 & 10.09 & 29.12 & 44.20 & 24.52 \\
 & RS-M-CLIP & ViT-B & Cap-5 / 1 Translation & 65.49 & 77.66 & 81.42 & 59.47 & 75.80 & 82.74 & 73.76 \\ \hline
\multirow{2}{*}{Korean} & CLIP-XLM-RoBERTa & ViT-B & Zero-shot & 7.52 & 20.13 & 28.32 & 8.45 & 24.65 & 37.30 & 21.06 \\
 & RS-M-CLIP & ViT-B & Cap-5 / 1 Translation & 65.71 & 76.33 & 81.86 & 58.32 & 74.38 & 81.20 & 72.97 \\ \hline
\multirow{2}{*}{Chinese} & CLIP-XLM-RoBERTa & ViT-B & Zero-shot & 9.29 & 20.35 & 29.87 & 9.51 & 25.35 & 40.22 & 22.43 \\
 & RS-M-CLIP & ViT-B & Cap-5 / 1 Translation & 64.38 & 77.21 & 82.52 & 57.39 & 75.40 & 81.64 & 73.09 \\ \hline
\multirow{2}{*}{Russian} & CLIP-XLM-RoBERTa & ViT-B & Zero-shot & 9.07 & 22.57 & 32.97 & 9.25 & 27.12 & 41.64 & 23.77 \\
 & RS-M-CLIP & ViT-B & Cap-5 / 1 Translation & 67.48 & 76.99 & 82.52 & 58.89 & 75.09 & 81.90 & 73.81 \\ \hline
\end{tabular}
}
\end{table*}

\vspace{-0.15cm}
\section{Experiments}

This section presents experimental results, assessing baselines against fine-tuned CLIP models in different tasks and settings.

\vspace{-0.15cm}
\subsection{Fine-Tuning with Contrastive Learning}

Table~\ref{table:clip-objective} reports the zero-shot performance of generalist CLIP models with ViT-B and ViT-L image encoders, in the cross-modal retrieval task and when considering different English datasets. We observe a reasonable performance, with only a slightly lower mean recall than baseline models specifically developed for the task, such as AMFMN, GaLR, and KCR. As expected, a larger CLIP image encoder improves the performance across all the datasets. 

Previous work had already observed that fine-tuning CLIP to the remote sensing domain leads to improved retrieval results, contributing with different CLIP models such as CLIP-RSICD~\cite{pal2021Fine}, GeoRSCLIP~\cite{zhang2024rs5m}, and RemoteCLIP~\cite{remoteclip}. Authors either used publicly available datasets or developed pipelines to create image-caption pairs for training. We aggregated 5 different English image-captioning remote sensing datasets, i.e. the collection which we named Cap-5, and fine-tuned a CLIP model with a ViT-L vision encoder, comparing its results with the RemoteCLIP model of the same size, which is the current state-of-the-art. While using  only 30\% of the size of the RemoteCLIP dataset, our English-only CLIP/Cap-5 model has a significantly better performance in the RSITMD (62.26\% vs 50.52\% mR) dataset, and a competitive performance in the RSICD (34.61\% vs 36.35\% mR) and UCM datasets (55.75\% vs 54.68\% mR). This can be explained by the characteristics of the instances in the RemoteCLIP data collection, which besides the standard benchmark datasets also includes instances derived from object detection and semantic segmentation datasets, with a high image resolution and domain diversity. While covering a broad set of conditions, the synthetic data can also be associated to a distribution shift regarding the datasets used for retrieval evaluation. 

CLIP/Cap-5 also obtains higher results than KTIR~\cite{mi2024knowledge} in RSITMD (62.26\% vs 50.52\% mR), but KTIR has a wide gap to it in RSICD (34.61\% vs 45.21\% mR) and UCM (55.75\% vs 58.76\% mR).

\vspace{-0.15cm}
\subsection{Comparing Multilingual Encoders}

We compared the original OpenAI CLIP model to a similar CLIP model trained with a collection of instances from LAION-5B~\cite{schuhmann2022laion}, specifically a multilingual model trained on 100 different languages, with a ViT-B image encoder and a XLM-RoBERTa~\cite{conneau2019unsupervised} text encoder. We refer to this model as CLIP-XLM-RoBERTa. 

As can be seen in Table~\ref{table:clip-objective}, the zero-shot performance of CLIP-XLM-RoBERTa surpasses that of the OpenAI CLIP models, across the different English datasets. Even when compared to the model that features a larger vision encoder, its performance is higher in RSICD (18.69\% vs 17.56\% mR), similar in RSITMD (27.46\% vs 27.88\% mR), and only worse in UCM (36.48\% vs 41.76\% mR). These results suggest clear benefits in using CLIP models pre-trained on larger datasets, even if the larger datasets are multilingual. 

We also fine-tuned CLIP-XLM-RoBERTa with Cap-5, and observed significant improvements. Compared with our previous larger model (i.e., CLIP-L) that was also fine-tuned with Cap-5, a large improvement can be observed in RSICD (45.56\% vs 34.61\% mR), and similar results are seen in UCM (55.13\% vs 55.75\% mR) and RSITMD (61.42\% vs 62.26\% mR). Compared with the RemoteCLIP model of larger size, a big improvement can be seen in RSITMD (61.42\% vs 50.52\% mR), and competitive performance is observed in RSICD (45.56\% vs 36.35\% mR) and UCM (55.13\% vs 54.68\% mR). 

CLIP-XLM-RoBERTa fine-tuned with Cap-5 is also better than KTIR in RSICD (45.56\% vs 45.1\% mR) and RSITMD (61.42\% vs 54.27\% mR), although worse in UCM (55.13\% vs 58.76\% mR). 

Together, our experimental results consolidate the improvements gained from the pre-trained weights of CLIP-XLM-RoBERTa, and the suitability of contrastive learning to adapt CLIP to the remote sensing domain. Considering that the Cap-5 dataset is a suitable source for training CLIP models, we then analyzed other training mechanisms besides the standard contrastive objective.

\vspace{-0.15cm}
\subsection{Self-Distillation and Contrastive Learning}

\begin{figure*}[t!]
\centering
\includegraphics[width=\textwidth]{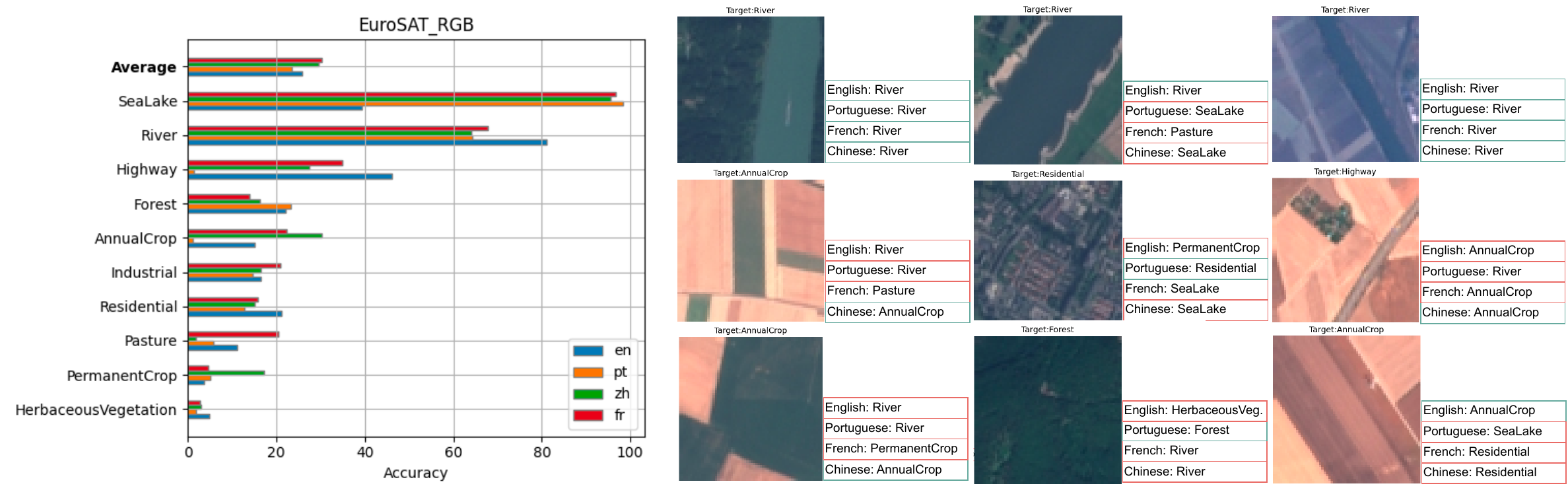}
\caption{Zero-shot classification performance on the EuroSAT dataset, across different languages.} \label{fig:zc-eurosat}
\end{figure*}

Results for our proposed model, which combines contrastive learning with a self-distillation objective, can also be seen in Table~\ref{table:clip-objective}. A large improvement is achieved in the RSICD and RSITMD datasets, even over encoders of larger size. The mean recall reaches an impressive value of 61.64\% in RSICD (+16.43\% over the SOTA model, i.e. KTIR) and 71.36\% in RSITMD (+17.09\% over KTIR). However, in the UCM dataset, the results are worse (-11.38\% over KTIR). This might suggest that the lower proportion of the UCM dataset instances within the Cap-5 dataset (approx. 4\%), compared to RSICD (approx. 22\%) and RSITMD (approx. 9.5\%), might lead the model to optimize more to characteristics of the datasets with larger size, hindering performance on UCM.

Detailed experiments were also conducted considering captions translated into different languages. For each caption in English within Cap-5, a translation is obtained in 9 other different languages (i.e., Portuguese, Spanish, French, German, Dutch, Italian, Korean, Chinese, and Russian). During training, for each image at each epoch, one of the possible translations of a caption is sampled, thus performing augmentations in the text, while still maintaining the total volume of image-caption pairs seen during training. This is the main RS-M-CLIP model and corresponds to the lines shown in red in the results presented in Table~\ref{table:clip-objective}. The performance for the image-text retrieval task can be further improved with this strategy,  across all English datasets (+2.07\% in RSICD, +2.06\% in UCM, and +3,19\% in RSITMD). Even in UCM, where our method had lower scores, the results were improved, particularly for the text-to-image direction. These findings confirm that caption translation is a form of text data augmentation that can significantly improve the retrieval performance~\cite{fan2024improving,vasu2024clip,li2024scaling}. 

\vspace{-0.15cm}
\subsection{Zero-shot Image Classification}

We also evaluated zero-shot classification through 12 image classification datasets, comparing our method against other CLIP-based approaches. The datasets used for evaluation are RSI-CB128~\cite{li2020rsi}, RSI-CB256~\cite{li2020rsi}, WHU-earth~\cite{zhao2015dirichlet}, EuroSAT~\cite{helber2019eurosat}, MLRSNet~\cite{qi2020mlrsnet}, PatternNet~\cite{zhou2018patternnet}, RESISC45~\cite{cheng2017remote}, AID~\cite{xia2017aid}, RSSCN7~\cite{zou2015deep}, OPTIMAL-31~\cite{wang2018scene}, RSC11~\cite{zhao2016feature}, and WHU-RS19~\cite{xia2010structural}. The results are compiled in Table~\ref{table:classification}, which presents accuracy values.

Overall, CLIP-XLM-RoBERTa has a higher average accuracy compared to the original OpenAI model, and is competitive with the OpenAI model featuring the larger vision encoder. The main RS-M-CLIP model, trained with translated data as augmentations, generally surpasses the RemoteCLIP model of the same size (i.e., better results are obtained in 8 out of 12 datasets), and is on average slightly lower than the RemoteCLIP model with a larger size, surpassing it in 3 out of the 12 datasets.

\vspace{-0.15cm}
\subsection{Multilingual Retrieval}

We also evaluated the multilingual capabilities of the main RS-M-CLIP model, considering image-text retrieval across different languages. Since there are no established multilingual benchmarks, the same translation transformation used for the training data was applied to the test splits of RSICD and RSITMD. The results reported by Rahhal et al.~\cite{rahhal2022multilanguage} are included as a comparison, since this is the only other study that evaluated cross-modal image-text retrieval in a multilingual scenario. However, since the data used for evaluation is not the same (i.e., different translations were used), the results cannot be directly compared. We evaluated retrieval over the 10 languages for which we obtained translations, including Portuguese, Spanish, French, German, Dutch, Italian, Korean, Chinese, and Russian. Of these, French and Italian are also considered by Rahhal et al.~\cite{rahhal2022multilanguage}. Results are compiled in Tables~\ref{table:multilingual-rsicd} and~\ref{table:multilingual-rsitmd}. 

The results show that, for the RSICD and RSITMD datasets, our method obtains much higher results compared to those reported by Rahhal et al.~\cite{rahhal2022multilanguage}. Across languages, it can be seen that the performance is very similar to the original results obtained in English.  

Interestingly, lower results can be observed for the German language, across the different retrieval datasets. This can be explained either by a lower performance of the original CLIP-XLM-RoBERTa model for the German language, or a lower quality of the translation data obtained for German from the translation model.

\begin{figure}[t!]
\centering
\includegraphics[width=\linewidth]{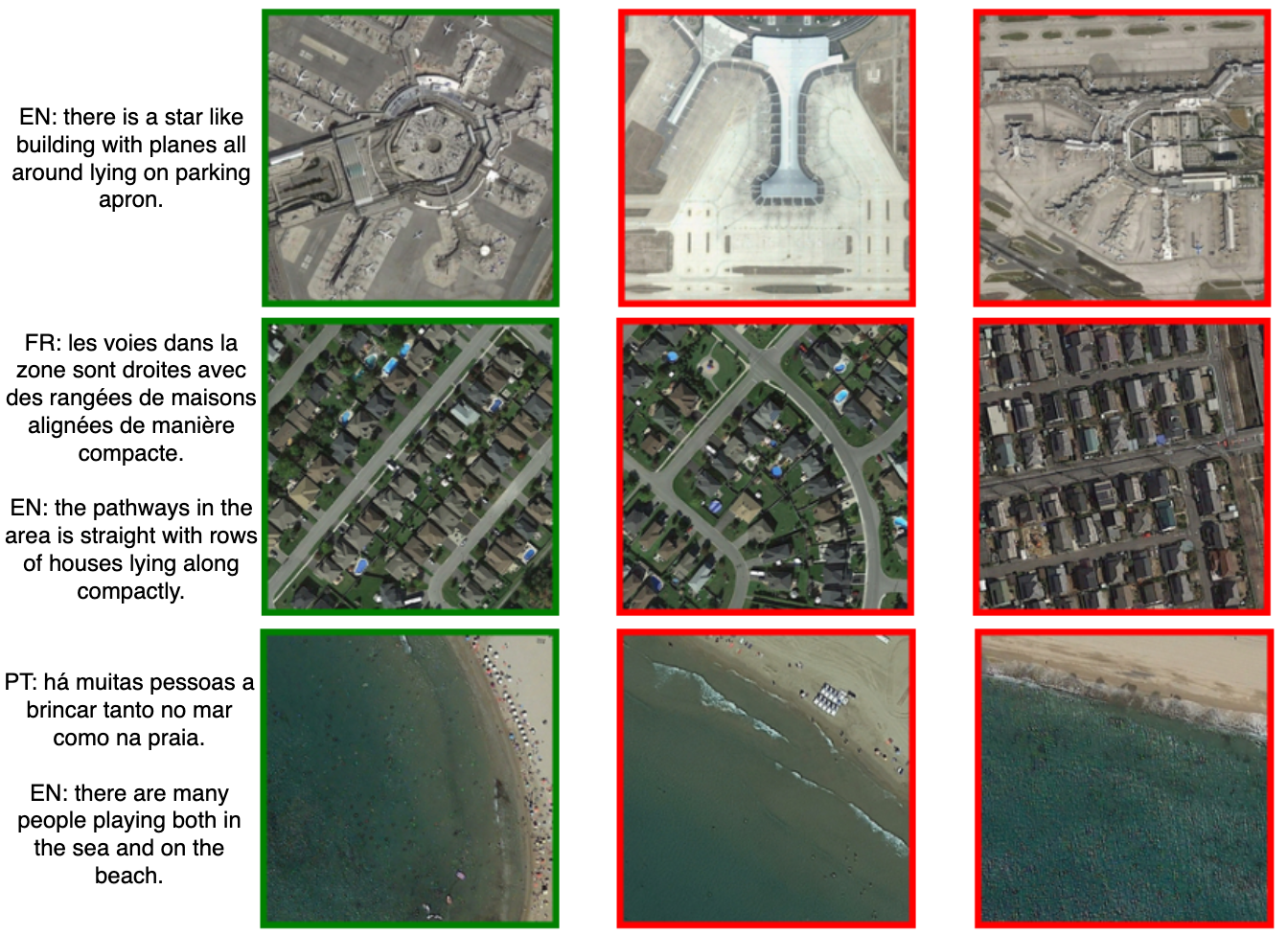}
\vspace{-0.15cm}
\caption{Examples for image retrieval, ordered by similarity, given different captions in English, Portuguese, and French.} \label{fig:multi-ret}
\end{figure}

\begin{figure}[t!]
    \centering
    \centering
    \includegraphics[width=\linewidth]{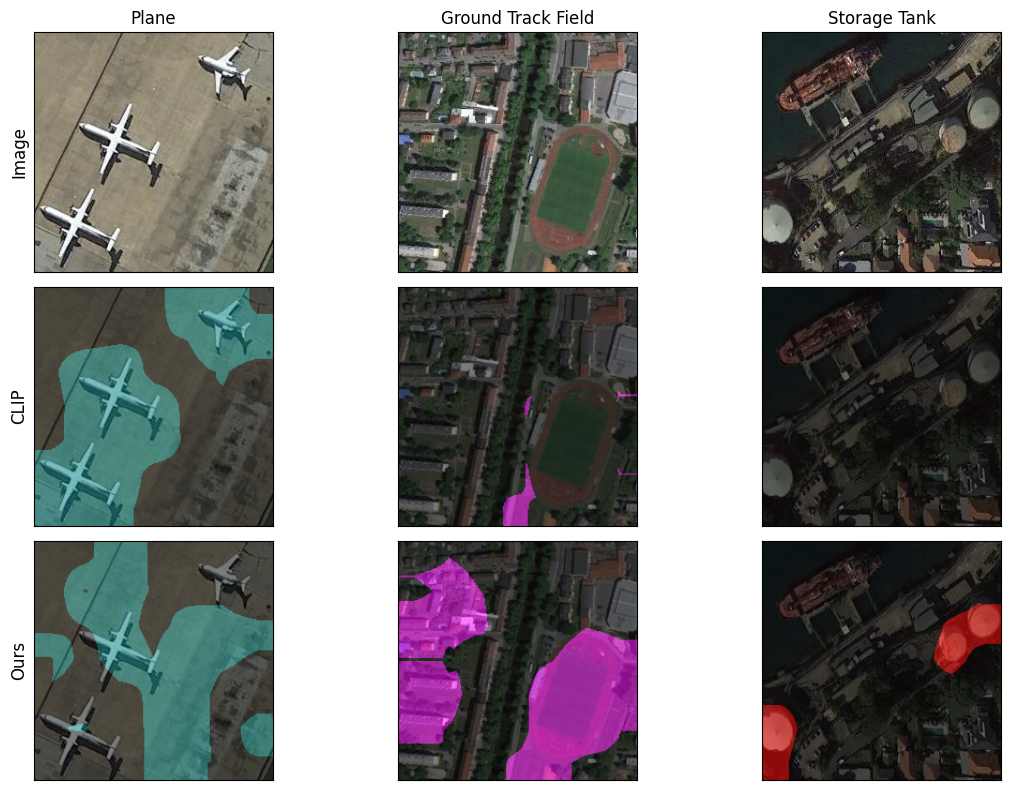}
    \vspace{-0.15cm}    
    \caption{Semantic masks obtained with the NACLIP~\cite{hajimiri2024pay} method from RS-M-CLIP and CLIP, across different scenes.} \label{fig:masks}
\end{figure}

\vspace{-0.15cm}
\section{Qualitative Analysis}

The results of an experiment illustrating the multilingual capabilities of RS-M-CLIP can be visualized in Figure~\ref{fig:zc-eurosat}. In this test, we took the images from the EuroSAT dataset to assess zero-shot classification performance, and we considered class names in 4 languages: English, Portuguese, Chinese, and French. For every language, the prompt for zero-shot classification was translated together with the class names. It can be seen that the image classification performance, on average, is similar across the languages, with French and Chinese even surpassing the English results. On the right part of Figure~\ref{fig:zc-eurosat}, we show specific image examples.

The multilingual retrieval capabilities are also illustrated in Figure~\ref{fig:multi-ret}, with examples of textual queries in English, Portuguese, and French. Images that correctly match the text are marked with green edges, while false positives are marked with red edges. From left to right, the three images for each query are ranked by the CLIP similarity. It can be seen that the top results returned by RS-M-CLIP are still closely related to the input textual query and, in all 3 cases, the correct image was returned in the first position for both the English and non-English descriptions.

Finally, we provide a visualization of semantic segmentation masks obtained from RS-M-CLIP, comparing them against masks derived from the original OpenAI CLIP model in Figure~\ref{fig:masks}. The masks are obtained with the NACLIP method~\cite{hajimiri2024pay}, an approach for open-vocabulary semantic segmentation that does not require training any new parameters. Instead, NACLIP modifies the CLIP architecture to improve the segmentation performance by increasing attention values between neighboring patches, based on the assumption that adjacent patches tend to belong to the same class. This is achieved by introducing a Gaussian kernel over the attention mechanism, and by eliminating components of the final encoder layer of the vision transformer, such as the feedforward layer and residual skip-connections. 
The examples in Figure~\ref{fig:masks} are from the NWPU-Captions dataset, considering three different classes present in the iSAID segmentation dataset~\cite{waqas2019isaid}, which features 15 different classes in total. 
The figure plots segmentation masks with classes related to the image contents, in particular for the classes \texttt{plane}, \texttt{ground track field}, and \texttt{storage tank}. 
To our surprise, the semantic masks obtained by CLIP are reasonable across different scenes. Still, RS-M-CLIP can more accurately describe most classes, as shown for the ground track field in the middle images, and the storage tanks in the images in the right.

\vspace{-0.15cm}
\section{Conclusions and Future Work}\label{end}

This work proposed RS-M-CLIP, i.e. a novel multilingual CLIP model for the remote sensing domain, that was trained on a mixture of English and machine translated data by combining a self-distillation objective with standard contrastive learning. RS-M-CLIP achieves impressive results in tasks such as cross-modal retrieval, and it also supports multilingual inputs. 

For future work, it would be interesting to study approaches for improving the compositional reasoning abilities of CLIP in the remote sensing domain~\cite{yuksekgonul2022and}, e.g. further extending RS-M-CLIP through training with hard negative examples. The training methodology can also be further extended for improving fine-grained visual abilities~\cite{wang2024diffusion,Monsefi2024DetailCLIP}. RS-M-CLIP can also be integrated with large language models, enabling tasks that involve language generation conditioned on remote sensing imagery~\cite{silva2024large,zhan2024skyeyegpt}. Finally, it would also be interesting to adapt/extend RS-M-CLIP to other remote sensing  modalities, going beyond the use of RGB images~\cite{zavras2024mind}.

\vspace{-0.15cm}
\begin{acks}
This research was supported by the Portuguese Recovery and Resilience Plan through project C645008882-00000055 (i.e., the Center For Responsible AI), and also by the Fundação para a Ciência e Tecnologia (FCT), specifically through the project with reference UIDB/50021/2020 (DOI: 10.54499/UIDB/50021/2020), and the project with reference UIDP/04516/2020 (DOI: 10.54499/UIDB/04516/2020).
\end{acks}

\bibliographystyle{ACM-Reference-Format}
\bibliography{tidy}


\appendix

\section{Generating Textual Captions from Retrieved Examples}
\label{appendix:captioning}

\begin{figure}[h]
\centering
\includegraphics[width=\columnwidth]{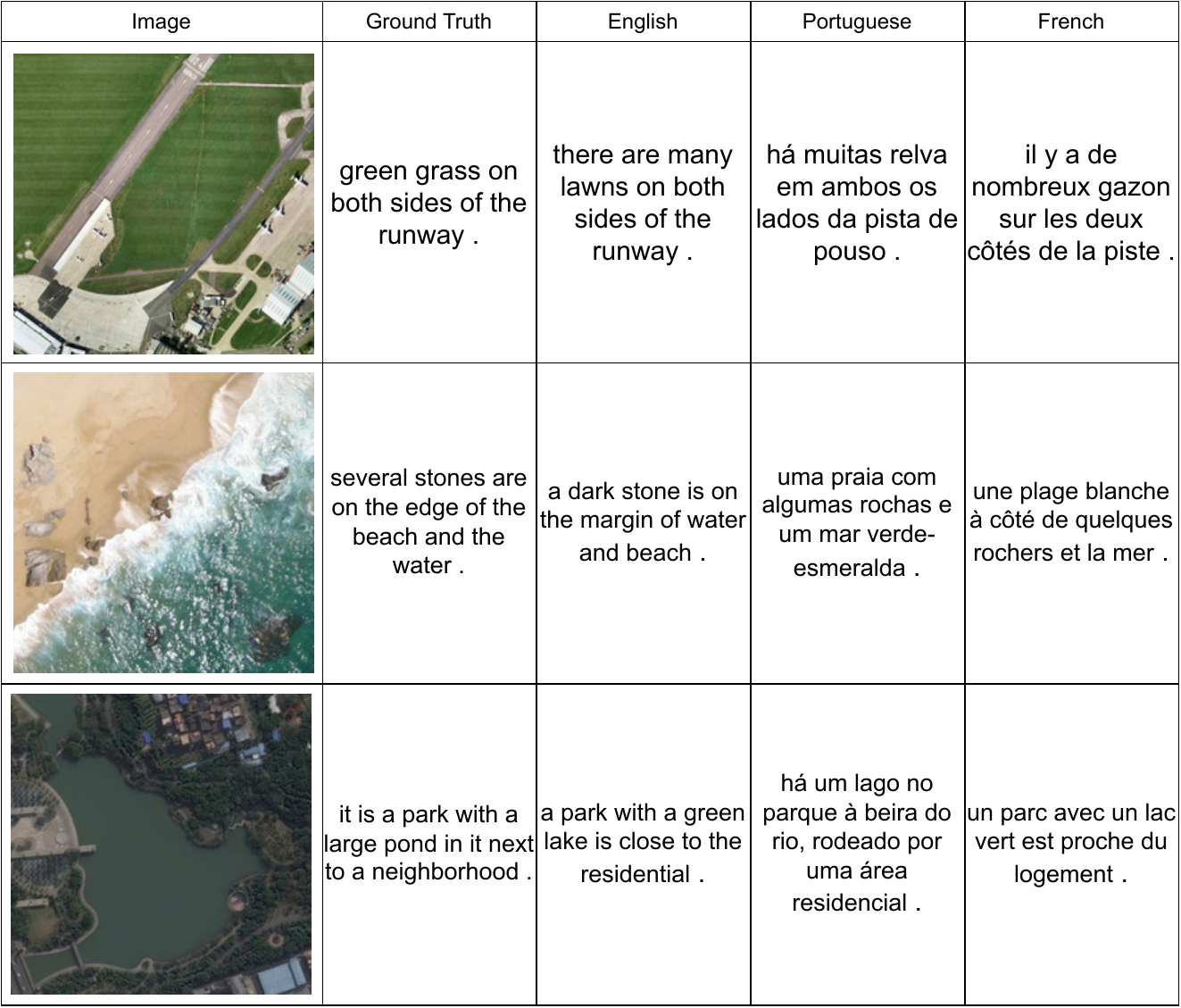}
\caption{Examples of captions generated with an approach based on LMCap~\cite{ramos2023lmcap}, using an LLM for generating captions conditioned on retrieved examples obtained by RS-M-CLIP.} \label{fig:lmcap}
\end{figure}

\begin{table*}[t!]
\centering
\caption{LMCap image captioning results, obtained by retrieving similar captions to the image, either in English or in the target language, and asking the language model to generate a caption in the target language. 
}
\label{table:lmcap-prompt1}
\vspace{-0.25cm}
\begin{tabular}{ccccccccccccc}
\hline
\multirow{2}{*}{Language} & \multicolumn{3}{c}{RSICD} & \multicolumn{3}{c}{UCM-Captions} & \multicolumn{3}{c}{Sydney-Captions} & \multicolumn{3}{c}{NWPU-Captions} \\ \cline{2-13} 
 & BLEU\_1 & BLEU\_4 & CIDEr & BLEU\_1 & BLEU\_4 & CIDEr & BLEU\_1 & BLEU\_4 & CIDEr & BLEU\_1 & BLEU\_4 & CIDEr \\ \hline
\multicolumn{13}{c}{\textbf{Supervised Methods for the English language}} \\ \hline
MLCA-NET~\cite{cheng2022NWPUCaptions} & 0.757 & 0.461 & 2.356 & 0.826 & 0.668 & 3.240 & 0.831 & 0.580 & 2.324 & 0.745 & 0.478 & 1.164 \\
RSGPT~\cite{hu2023rsgpt} & 0.703 & 0.368 & 1.029 & 0.861 & 0.657 & 3.332 & 0.823 & 0.622 & 2.731 & - & - & - \\
SkyEyeGPT~\cite{zhan2024skyeyegpt} & 0.867 & 0.600 & 0.837 & 0.907 & 0.784 & 2.368 & 0.919 & 0.774 & 1.811 & - & - & - \\ \hline \hline
\multicolumn{13}{c}{\textbf{Examples in English and Generated Caption in the Target Language}} \\ \hline
\cellcolor{red!15}English & \cellcolor{red!15}0.552 & \cellcolor{red!15}0.163 & \cellcolor{red!15}0.617 & \cellcolor{red!15}0.776 & \cellcolor{red!15}0.558 & \cellcolor{red!15}2.742 & \cellcolor{red!15}0.707 & \cellcolor{red!15}0.480 & \cellcolor{red!15}2.028 & \cellcolor{red!15}0.744 & \cellcolor{red!15}0.466 & \cellcolor{red!15}1.111 \\
Portuguese & 0.484 & 0.140 & 0.489 & 0.680 & 0.398 & 1.805 & 0.661 & 0.387 & 1.293 & 0.677 & 0.394 & 0.894 \\
Spanish & 0.518 & 0.157 & 0.523 & 0.681 & 0.406 & 1.835 & 0.641 & 0.387 & 1.361 & 0.699 & 0.406 & 0.944 \\
French & 0.492 & 0.151 & 0.493 & 0.682 & 0.396 & 1.475 & 0.644 & 0.384 & 1.178 & 0.691 & 0.402 & 0.859 \\
German & 0.406 & 0.104 & 0.319 & 0.619 & 0.320 & 1.226 & 0.597 & 0.305 & 1.090 & 0.643 & 0.341 & 0.688 \\
Dutch & 0.526 & 0.137 & 0.476 & 0.709 & 0.384 & 1.589 & 0.678 & 0.395 & 1.315 & 0.697 & 0.373 & 0.836 \\
Italian & 0.485 & 0.147 & 0.508 & 0.677 & 0.403 & 1.789 & 0.578 & 0.301 & 0.937 & 0.661 & 0.374 & 0.818 \\
Korean & 0.524 & 0.155 & 0.456 & 0.641 & 0.337 & 1.115 & 0.617 & 0.307 & 0.780 & 0.717 & 0.372 & 0.801 \\
Chinese & 0.603 & 0.213 & 0.655 & 0.770 & 0.475 & 1.697 & 0.732 & 0.427 & 1.053 & 0.752 & 0.429 & 1.002 \\
Russian & 0.318 & 0.069 & 0.243 & 0.561 & 0.253 & 0.897 & 0.498 & 0.195 & 0.601 & 0.566 & 0.293 & 0.574 \\ \hline
Average & 0.471 & 0.133 & 0.454 & 0.665 & 0.397 & 1.570 & 0.617 & 0.340 & 1.133 & 0.665 & 0.367 & 0.820 \\ \hline \hline
\multicolumn{13}{c}{\textbf{Examples and Generated Caption in the Target Language}} \\ \hline
\cellcolor{red!15}English & \cellcolor{red!15}0.552 & \cellcolor{red!15}0.163 & \cellcolor{red!15}0.617 & \cellcolor{red!15}0.776 & \cellcolor{red!15}0.558 & \cellcolor{red!15}2.742 & \cellcolor{red!15}0.707 & \cellcolor{red!15}0.480 & \cellcolor{red!15}2.028 & \cellcolor{red!15}0.744 & \cellcolor{red!15}0.466 & \cellcolor{red!15}1.111 \\
Portuguese & 0.485 & 0.154 & 0.553 & 0.727 & 0.485 & 2.219 & 0.678 & 0.423 & 1.414 & 0.680 & 0.397 & 0.901 \\
Spanish & 0.528 & 0.168 & 0.573 & 0.723 & 0.470 & 2.238 & 0.677 & 0.416 & 1.438 & 0.701 & 0.407 & 0.955 \\
French & 0.509 & 0.171 & 0.549 & 0.717 & 0.458 & 1.945 & 0.615 & 0.386 & 1.287 & 0.689 & 0.397 & 0.849 \\
German & 0.432 & 0.115 & 0.375 & 0.702 & 0.437 & 1.734 & 0.617 & 0.344 & 1.435 & 0.645 & 0.341 & 0.696 \\
Dutch & 0.527 & 0.151 & 0.519 & 0.717 & 0.433 & 1.806 & 0.691 & 0.393 & 1.358 & 0.699 & 0.372 & 0.841 \\
Italian & 0.489 & 0.157 & 0.543 & 0.737 & 0.528 & 2.358 & 0.625 & 0.377 & 1.081 & 0.662 & 0.377 & 0.827 \\
Korean & 0.533 & 0.166 & 0.447 & 0.715 & 0.458 & 1.540 & 0.611 & 0.310 & 0.857 & 0.742 & 0.393 & 0.863 \\
Chinese & 0.582 & 0.207 & 0.640 & 0.754 & 0.499 & 1.792 & 0.708 & 0.413 & 1.111 & 0.748 & 0.427 & 0.993 \\
Russian & 0.337 & 0.085 & 0.275 & 0.646 & 0.409 & 1.438 & 0.561 & 0.277 & 0.967 & 0.568 & 0.292 & 0.570 \\ \hline
Average & 0.475 & 0.142 & 0.487 & 0.707 & 0.458 & 1.929 & 0.631 & 0.368 & 0.274 & 0.668 & 0.368 & 0.824 \\ \hline \hline
\end{tabular}
\vspace{-0.25cm}
\end{table*}

Although the proposed RS-M-CLIP model consists only of encoders, the generation of image captions can also be performed by integrating it with a language model decoder. We illustrate these capabilities, following a framework similar to LMCap~\cite{ramos2023lmcap}, where captions with similar CLIP embeddings are retrieved for a given input image and used to prompt a Large Language Model (LLM) to generate a new caption, without receiving image features. 

We used RS-M-CLIP in this setup to generate captions for different remote sensing image captioning datasets. Considering the multilingual abilities of RS-M-CLIP, we evaluated image captioning in different languages other than English, with two setups: (i) retrieving captions in English and prompting the LLM to generate a caption in the target language, and (ii) retrieving captions already in the target language and asking the LLM to generate a caption. Both setups used the aforementioned TowerInstruct LLM~\cite{alves2024tower}. 

The prompt for multilingual image captioning is as follows, considering $K$ retrieved captions as examples describing the image, and with translated captions into the target language in the case of scenario (ii): \texttt{You are an intelligent image captioning bot tasked with describing remote sensing images. Similar images have the following captions: <caption~$1$>, ..., <caption~$K$>. A creative short caption that can describe this image in <language> is:}

The large language model is also prompted with a fixed set of $N$-shot examples of multiple image classes from the RSICD dataset, with $K$ retrieved captions per example and with the output caption in the target language. The involved parameters were set to $N=6$ and $K=4$, where the $N$ examples are each represented through the same textual prompt that was described above, and selected from different classes in the RSICD dataset (i.e., \texttt{airport, denseresidential, baseballfield, parking, stadium,} and \texttt{playground}). We found that the fixed examples could effectively be used across the different datasets, as evidenced by high captioning quality scores, particularly for English. In the second setup 
the entire prompt was also translated accordingly.

The results of the first setup can be seen in Table~\ref{table:lmcap-prompt1}, with scores from previous English-only supervised methods also included for comparison. Performance was evaluated with standard metrics from the \texttt{COCOeval} package\footnote{\url{https://github.com/salaniz/pycocoevalcap}} with SacreBLEU tokenization\footnote{\url{https://github.com/mjpost/sacrebleu}}, namely BLEU-1, BLEU-4, and CIDER. It can be seen that, for the English language, the proposed approach obtains reasonable performance across the datasets, and is very similar to MLCA-Net~\cite{cheng2022NWPUCaptions} for the NWPU-Captions dataset. Results are also consistent across languages. In turn, the results of the second setup, given at the bottom of Table~\ref{table:lmcap-prompt1}, show that the image captioning performance is generally improved across languages, confirming the strong multilingual abilities of RS-M-CLIP. The setup with retrieved examples already in the target language helps the LLM to generate better captions.

Figure~\ref{fig:lmcap} shows some examples from the results of these experiments, for the English, Portuguese, and French languages. 
In the first row, a consistent description of the airport is given across all three languages. Regarding the image with a beach, the descriptions differ slightly, in particular emphasizing differently the colors of objects of interest (i.e., the color of the stone in English, while the color of the sea is mentioned in Portuguese, and the color of the beach in French). Lastly, a similar description is generated across all three languages for the third example image.

\end{document}